\documentclass{article} % For LaTeX2e

\usepackage[final]{colm2026_conference}

\usepackage[T1]{fontenc}
\usepackage{lineno}

\usepackage{tabularx}

\usepackage{microtype}
\usepackage{graphicx}
\usepackage{subcaption}
\usepackage{booktabs}
\usepackage{array}
\usepackage{rotating}
\usepackage{enumitem}
\usepackage{float}
\usepackage{tikz}
\usetikzlibrary{arrows.meta, positioning, calc}

\usepackage{hyperref}
\usepackage{url}
\definecolor{darkblue}{rgb}{0, 0, 0.5}
\hypersetup{colorlinks=true, citecolor=darkblue, linkcolor=darkblue, urlcolor=darkblue}

\usepackage{algorithm}
\usepackage{algorithmic}
\usepackage{fvextra}
\usepackage[most]{tcolorbox}
\usepackage{xcolor}
\usepackage{wrapfig}

\definecolor{promptbg}{RGB}{248,248,248}
\definecolor{promptframe}{RGB}{170,170,170}

\newtcblisting{PromptBlock}[2][]{
  listing only,
  breakable,
  enhanced,
  colback=promptbg,
  colframe=promptframe,
  boxrule=0.35pt,
  arc=1mm,
  left=0.5mm,
  right=0.5mm,
  top=0.5mm,
  bottom=0.5mm,
  boxsep=0.5mm,
  before skip=0.5em,
  after skip=0.2em,
  listing options={
    breaklines=true,
    basicstyle=\fontsize{5.5}{5.9}\selectfont\ttfamily,
    tabsize=2,
    showspaces=false,
    showstringspaces=false
  },
  title={#2},
  label={#1}
}

\usepackage{amsmath}
\usepackage{amssymb}
\usepackage{mathtools}
\usepackage{amsthm}

\usepackage[capitalize,noabbrev]{cleveref}

\theoremstyle{plain}

\theoremstyle{definition}

\theoremstyle{remark}

\title{Test-Time Self-Play: Scaling without a Verifier
}

\author{Akira Yoshiyama\thanks{Equal contribution. Correspondence to: \texttt{akira@tufalabs.ai}.} \\
ETH Zurich \& Tufa Labs \\
\And
Akmal Ashirmatov\footnotemark[1] \\
ETH Zurich \\
\And
Jonas H\"ubotter \\
ETH Zurich \\
\AND
\rlap{Ido Hakimi}\hphantom{Akira Yoshiyama\textsuperscript{*}} \\
\rlap{ETH Zurich}\hphantom{ETH Zurich \& Tufa Labs}\hspace{25.9pt} \\
\And
\rlap{Andreas Krause}\hphantom{Akmal Ashirmatov\textsuperscript{*}} \\
ETH Zurich \\
\And
\hphantom{Jonas H\"ubotter} \\ % invisible third column: row 2 mirrors row 1 so Andreas sits under Akmal
\hphantom{ETH Zurich} \\
}

\newif\ifmainbodydisplays
\makeatletter
\g@addto@macro\normalsize{\ifmainbodydisplays
  \abovedisplayskip 10pt plus 3pt minus 2pt\belowdisplayskip\abovedisplayskip
  \abovedisplayshortskip\abovedisplayskip\belowdisplayshortskip\abovedisplayskip\fi}
\makeatother
\newskip\savedADS \newskip\savedBDS \newskip\savedADSS \newskip\savedBDSS

\begin{document}
\savedADS=\abovedisplayskip \savedBDS=\belowdisplayskip \savedADSS=\abovedisplayshortskip \savedBDSS=\belowdisplayshortskip
\mainbodydisplaystrue\normalsize

\ifcolmsubmission
\linenumbers
\fi

\maketitle
\setlength{\textfloatsep}{12pt plus 2pt minus 2pt}% main-body: tighter gap below top-of-page figures
\captionsetup[figure]{skip=6pt}% main-body figures: half the default plot-to-caption gap

\begin{abstract}
Test-time scaling methods spend additional compute at deployment to improve a large language model's performance on a hard target problem. Most methods rely on a verifier that scores candidate solutions for correctness, using it to select among or refine the model's outputs. In many deployment regimes, however, a strong verifier on the target problem is unavailable. We introduce \textbf{Test-Time Self-Play (TTSP)}, an approach to test-time scaling using asymmetric self-play without reliance on a verifier. On a difficulty-calibrated subset of LiveCodeBench v6, TTSP is the best-performing method at every verifier strength we evaluate, and the gap to the strongest baseline widens as the verifier weakens; in the no-verifier limit, TTSP yields a 35\% absolute improvement in mean@8. The gains hold across a range of verifier strengths, except when the base policy has near-zero coverage of the correct answer.

\end{abstract}

\section{Introduction}
\label{sec:intro}

\begin{wrapfigure}{r}{0.5\textwidth}
\centering
\vspace{-14pt}
\includegraphics[width=\linewidth]{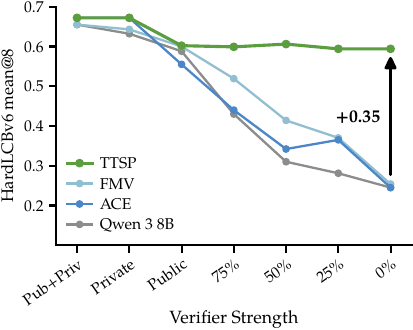}
\caption{\textbf{TTSP outperforms all baseline test-time scaling methods in the weak verifier regime.} TTSP and ACE are evaluated at converged checkpoints; FMV uses public and private input test cases.}
\label{fig:verifier_strength_hard}
\vspace{-20pt}
\end{wrapfigure}

How can a model improve on a single difficult target problem when it has no related training data and no reliable way to check its own answers? This question arises whenever a large language model's underlying knowledge is, in principle, sufficient to solve a problem, yet the model rarely, or never, generates the correct solution. Examples of this can be found across domains, including in competitive programming \citep{zheng2025livecodebenchpro}, theorem proving \citep{tsoukalas2024putnambench}, and molecule generation \citep{gao2022pmo}. % This question arises whenever a foundation model encounters a deployment context in which its underlying knowledge is, in principle, sufficient to answer the target problem, yet its single-shot accuracy is low: a freshly released competitive-programming problem where the algorithmic primitives are well within the model's capabilities, but a correct solution surfaces only on rare rollouts \citep{jain2024livecodebench}; a frontier theorem-proving target where the requisite lemmas are part of the model's training distribution, but the integrated proof essentially never assembles in one pass \citep{deepseek_prover_v2}; or a candidate molecule whose desirable bioactivity is implicitly recognized by the model's chemistry priors, but reliably surfaced only after extensive sampling \citep{thomas2025molexp}.

Test-time scaling aims to close this gap, spending additional compute at deployment to improve performance on the target problem. Many test-time scaling methods, such as rejection sampling~\citep{snell2024scaling}, rely on a verifier for the target problem that determines whether an attempt is correct. Similarly, prompt-optimization methods such as ACE \citep{zhang2026ace} that use execution feedback for evolving prompts. Other methods rely on auxiliary data, typically similar problems retrieved from a large corpus and used to train on a neighborhood of the target problem \citep{hubotter2025learningonjob, hubotter2025nearest}. Another set of methods requires neither verifier nor auxiliary training data, instead aggregating a signal from the model's own rollouts on the target problem through majority voting \citep{zuo2025ttrl, du2026dare, shafayat2025srt, launer2026fmv}.

We propose \textbf{Test-Time Self-Play (TTSP)}, a test-time scaling method that does not rely on auxiliary training data or strong verifiers. TTSP is inspired by training-time asymmetric self-play methods such as Absolute Zero \citep{zhao2025absolutezero} and R-Zero \citep{huang2025rzero}, which learn from self-generated synthetic problems alone. We adapt this approach to the single-target setting, concentrating the synthetic problems on a target problem rather than across a benchmark distribution. The only external requirement of TTSP is an interactive environment, such as a Python interpreter, to ground the model.

Our contributions are the following. \textbf{(i)} We define the test-time scaling setting with no auxiliary data and absent or weak verifiers, and characterize the behavior of existing test-time scaling methods. \textbf{(ii)} We propose TTSP, a test-time scaling method based on asymmetric self-play that is robust to weak verifiers.

\begin{figure}
\includegraphics[width=\textwidth]{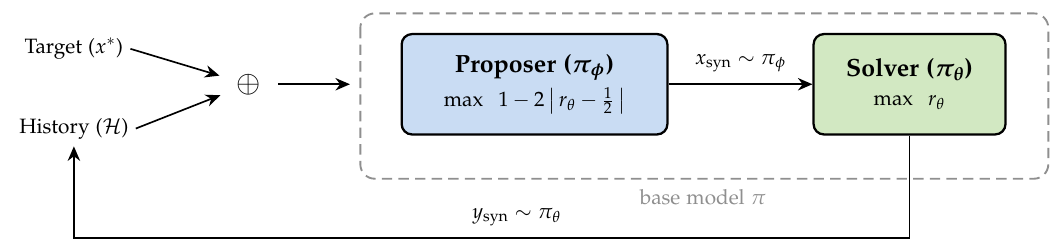}
\caption{\textbf{Overview of Test-Time Self-Play.}
Given a target problem $x^\star$ and previously generated synthetic problems, the proposer policy $\pi_\phi$ generates target-conditioned synthetic problems $\mathcal{D}$ consisting of a problem statement, tests, and a reference solution. The solver policy is rewarded for correctly solving the synthetic problems, while the proposer is rewarded for generating synthetic problems near the solver's current learning frontier (both via GRPO; see the red arrows). The synthetic problem history provides context for the proposer and fills a replay buffer $\mathcal{B}$. After a sufficient number of iterations, the test-time trained solver can then attempt the target problem.}
\label{fig:ttsp-overview}
\end{figure}

\section{Related Work}
\label{sec:related}

\paragraph{Test-time scaling.} Spending additional compute at deployment to improve performance on a target problem has been studied under the umbrellas of test-time scaling \citep{snell2024scaling, brown2024monkeys} and test-time training (TTT) \citep{sun2020ttt}. Sampling-based test-time scaling repeatedly draws rollouts from the policy and selects among them with a verifier or vote \citep{brown2024monkeys, launer2026fmv}. Weight-updating TTT specializes a model's weights to each target problem, with success across abstract reasoning \citep{akyurek2025surprising}, language modeling \citep{hardt2024ttt, hubotter2025nearest, sun2025lm}, and video generation \citep{dalal2025video}, using supervised losses on target problem neighborhoods drawn from external data \citep{hardt2024ttt, hubotter2025nearest, bertolissi2025, bagatella2025} or self-supervised losses over the input sequence \citep{sun2025lm, dalal2025video}. Most relevant to our work, \citet{zuo2025ttrl} extend TTT to RL on the test set using majority votes as pseudo-labels, while \citet{simonds2025ladder} generate problem variants at test-time and train via RL with numerical verification in symbolic integration. This connects to broader themes of unsupervised RL \citep{zhang2025unsup, agarwal2025entropy, zhou2025unsup, du2026dare} and self-improvement in LLMs \citep{zelikman2022star, gulcehre2023rest, lee2025selfimprovement}. How best to derive a strong learning signal for test-time scaling from the model alone, without recourse to external data or supervision, remains an open question.

\paragraph{Curriculum learning.} Introduced by \citet{bengio2009curriculum}, curriculum learning is the principle that ordering training examples from easy to hard accelerates optimization and improves generalization. It has since become a central design pattern in modern LLM training, underpinning pretraining and mid-training (DoReMi, \citealt{xie2023doremi}; OLMo 2, \citealt{walsh2025olmo}; Phi-4, \citealt{abdin2024phi4}) and, especially, RL-based reasoning post-training, where DeepSeek-R1 \citep{guo2025deepseekr1}, Kimi k1.5 \citep{kimi2025}, and adaptive schemes such as Self-Evolving Curriculum \citep{chen2025sec} and easy-to-hard generalization \citep{sun2024e2h} consistently outperform uniform sampling.

\paragraph{Self-play.} Self-play underlies many landmark results in reinforcement learning \citep{schmidhuber1991selfplay, tesauro1995tdgammon, silver2016alphago, silver2017alphazero, silver2018alphazero, berner2019dota}. Unlike fixed datasets, the self-play curriculum co-evolves with the learner and is in principle unbounded, a property recently argued to be central to continued progress beyond the limits of human-generated data \citep{silver2025era}. Recent work imports this property into LLM post-training by assigning a model dual proposer policy and solver policy roles with learning signal from automatic verifiers such as code execution, majority vote, or proof checkers \citep{zhao2025absolutezero, huang2025rzero, chen2025sqlm, kuba2025lsp, dong2025spiral}. However, the resulting curricula tend to concentrate on narrow regions of the task space, and gains from fully data-free self-play are often bounded by the base model's coverage \citep{chae2025coverage}. Prior self-play methods for LLMs operate at training time and target broad capability gains across a benchmark distribution. We instead apply self-play at \emph{test-time}, using target problem decomposition to build a specialized curriculum that concentrates adaptation on the target problem.

\paragraph{Context optimization.}
A complementary line of work adapts LLM behavior by modifying the context rather than the model weights. GEPA \citep{gepa2025} optimizes textual parameters through reflective prompt evolution, using verifier feedback to propose mutations and Pareto-style selection to retain complementary prompt variants. ACE \citep{zhang2026ace} extends this paradigm with a structured generator--reflector--curator decomposition: a generator attempts the task, a reflector extracts reusable lessons from execution feedback, and a curator incrementally updates a long-form playbook context while preserving useful details. This decomposition lets ACE accumulate more fine-grained domain knowledge than the prompt-level mutations of GEPA, and ACE reports consistent gains over GEPA across reasoning and tool-use benchmarks.

\section{Preliminaries}
\label{sec:problem}

We study test-time scaling with individual target problems. Let $x^\star$ denote a target problem and let $y$ denote a candidate solution. A policy $\pi$ generates rollouts $y \sim \pi(\cdot \mid x^\star)$. The oracle verifier,
\[
v^\star(y; x^\star) \in \{0,1\},
\]
indicates whether $y$ is a correct solution for $x^\star$. The oracle is used for evaluation, but is not assumed to be available for training or feedback at test-time. A deployment-time target verifier $v$ may be weak or absent. We further assume that no auxiliary training data (i.e., problems and their ground truth solutions $(x, y)$ related to $x^\star$) are available.
We ask whether a test-time scaling method can
be made robust in this setting.

We propose that this is possible by deriving a learning signal from \emph{synthetic problems} instead of directly from the target problem.
A \emph{synthetic problem} $\hat{x}$ is a self-contained  problem generated at test-time, conditioned on $x^\star$. We evaluate TTSP in the setting of competitive programming, in which $\hat{x}$ consists of a natural-language statement, an input/output specification, an example input/output pair, held-out input test cases, and a reference Python program. The reference program is not shown to the solver. Executing the reference solution on the held-out input test cases produces the ground truth outputs, which combined with the inputs define a synthetic verifier
\[
v_{\mathrm{syn}}(\hat{y}; \hat{x}) \in \{0,1\}
\]
The verifier identifies whether the attempted solution $\hat{y}$ passes all held-out test cases. We define synthetic problems to be self-contained training tasks with a deterministic verification mechanism. We do not require a formal decomposition relation between synthetic problems $\hat{x}$ and $x^\star$; the target problem dependence in TTSP is induced by conditioning the proposer on $x^\star$ and the synthetic problem history.

\begin{algorithm}[tb]
   \caption{Test-Time Self-Play}
   \label{alg:ttsp-detailed}
\begin{algorithmic}[1]
   \REQUIRE Target problem $x^\star$, proposer policy $\pi_\phi$, solver policy $\pi_\theta$, iterations $T$
   \STATE Initialize synthetic problem history $\mathcal{H} \gets \emptyset$
   \FOR{$t = 1$ {\bfseries to} $T$}
      \STATE $\mathcal{D}_t, \mathcal{B}_t, \mathcal{R}_\theta, \mathcal{R}_\phi \gets \emptyset$

      \STATE \textit{\textcolor{gray}{// Generate synthetic problems}}
      \STATE $\mathcal{C}_t \gets \textsc{ConstructContexts}(x^\star, \mathcal{H})$
      \FOR{each $c \in \mathcal{C}_t$}
         \STATE $\{\hat{x}^{(i)}\} \sim \pi_\phi(\cdot \mid c)$
         \STATE $\mathcal{A}, \mathcal{I} \gets \textsc{Filter}(\{\hat{x}^{(i)}\})$
         \STATE $\mathcal{D}_t \gets \mathcal{D}_t \cup \mathcal{A}$
         \STATE $\mathcal{R}_\phi \gets \mathcal{R}_\phi \cup \textsc{InvalidPenalty}(\mathcal{I})$
      \ENDFOR

      \STATE \textit{\textcolor{gray}{// Sample from history}}
      \STATE $\mathcal{B}_t \gets \textsc{SampleHistory}(\mathcal{H})$

      \STATE \textit{\textcolor{gray}{// Solver policy attempts synthetic problems}}
      \FOR{each $\hat{x} \in \mathcal{D}_t \cup \mathcal{B}_t$}
         \STATE $\{\hat{y}^{(j)}\} \sim \pi_\theta(\cdot \mid \hat{x})$
         \STATE $\{r_\theta^{(j)}\} \gets v_{\mathrm{syn}}(\{\hat{y}^{(j)}\}; \hat{x})$
         \STATE $\mathcal{R}_\theta \gets \mathcal{R}_\theta \cup \{(\hat{x}, \hat{y}^{(j)}, r_\theta^{(j)})\}$
         \STATE $\mathcal{H} \gets \textsc{UpdateHistory}(\mathcal{H}, \hat{x}, \{\hat{y}^{(j)}\}, \{r_\theta^{(j)}\})$
      \ENDFOR

      \STATE \textit{\textcolor{gray}{// Assign proposer policy rewards}}
      \STATE $\mathcal{R}_\phi \gets \mathcal{R}_\phi \cup \textsc{ProposerReward}(\mathcal{D}_t, \mathcal{R}_\theta)$

      \STATE \textit{\textcolor{gray}{// Update proposer policy and solver policy}}
      \STATE Update solver policy $\pi_\theta$ with GRPO using $\mathcal{R}_\theta$
      \STATE Update proposer policy $\pi_\phi$ with GRPO using $\mathcal{R}_\phi$
   \ENDFOR
   \STATE \textbf{return} adapted solver policy $\pi_\theta$
\end{algorithmic}
\end{algorithm}

\section{TTSP: Test-Time Self-Play}
\label{sec:method}

Test-Time Self-Play (TTSP) adapts a solver policy to a single target problem $x^\star$ by training on target-conditioned synthetic problems. A proposer policy generates synthetic problems, a solver policy attempts them, execution supplies rewards, and both policies are updated with GRPO. The target problem verifier is not used during TTSP training. Figure~\ref{fig:ttsp-overview} illustrates the  asymmetric self-play loop.

\paragraph{Self-play setup.}
TTSP uses two LoRA adapters over a shared frozen Qwen3-8B backbone \citep{hu2022lora,yang2025qwen3}: a proposer policy $\pi_\phi$ and a solver policy $\pi_\theta$. The proposer policy designs a synthetic curriculum conditioned on $x^\star$ and the training history, while the solver policy trains on the resulting synthetic problems. Separating the adapters gives each role its own reward signal and avoids the gradient conflicts we observed when using a single adapter for both roles. The memory cost remains close to a single 8B model, since the backbone is shared and only the adapters are updated. 

\paragraph{Algorithm overview.}
A step, $t$, of the TTSP algorithm begins with the construction of the proposer's prompt. The prompt (or context) contains a small sample of previously generated synthetic problems and the their solve rates (from the solver policy). The context also includes the target problem and instructions guiding the proposer to generate valid, relevant and appropriately challenging synthetic problems based on previous synthetic problems. The prompt in the first step does not contain previous synthetic problems.

The proposer then generates candidate synthetic problems and reference programs. Invalid candidates are removed by a domain-specific heuristic filter $\textsc{Filter}$.  The valid synthetic problems $\mathcal{D}_t$ are added to a buffer of old synthetic problems $\mathcal{B}_t$ which are sampled from the history of synthetic problems $\mathcal{H}$. Problems within the range of solve rates $[0.1, 0.8]$ are uniformly randomly sampled.

The solver policy then generates solutions for the combined set of synthetic problems $\mathcal{D}_t \cup \mathcal{B}_t$. The synthetic reward is calculated for each solution.

The history $\mathcal{H}$ is updated with the new synthetic problems, solver rollouts and rewards. The solver is then updated with GRPO using the synthetic rewards. The proposer is updated with GRPO using a transformation of the solver rewards $\textsc{ProposerReward}(\mathcal{D}_t, \mathcal{R}_\theta)$. The transformation is described later in this section.

After a sufficient number of steps, the aim is for the solver policy to have a more accurate distribution of solutions on the target problem $x^\star$. The solver can then be used to attempt the target problem.

Synthetic problems provide a dense training signal where direct training on $x^\star$ would be sparse, misleading (due to weak target problem verification) or unavailable (if no target problem verifier is available). TTSP produces many target-conditioned synthetic problems that are both at the edge of the solver's abilities (due to resampling of previous problems in the optimal difficulty range, and proposer reward transformation) and anchored to the target problem (due to the context construction). See Algorithm \ref{alg:ttsp-detailed} for the TTSP pseudocode.

\begin{wraptable}{r}{0.5\textwidth}
\vspace{-13pt}
\centering
\small
\setlength{\tabcolsep}{4pt}
\renewcommand{\arraystretch}{1.16}
\caption{\textbf{Example synthetic curriculum.}
The synthetic problems are generated by the proposer, teaching the solver concepts which are useful to recover the final strategy needed to solve the target problem.}
\label{tab:abc393e_ladder}
\begin{tabularx}{\linewidth}{@{}>{\hsize=1.13\hsize\raggedright\arraybackslash}X>{\hsize=0.87\hsize\raggedright\arraybackslash}X@{}}
\toprule
\addlinespace[0.2em]
\textbf{Target Problem $x^\star$} & \textbf{Concept} \\
\midrule
For each $A_i$, maximize the GCD of a size-$K$ subset containing $A_i$. &
Largest feasible divisor of $A_i$ \\
\midrule
\textbf{Synthetic Problems} & \textbf{} \\
\midrule
Find the largest integer $d$ that divides at least $K$ elements of the array. &
Feasibility predicate:
$\mathrm{cnt}[d] \ge K$ \\
\cmidrule[0.25pt](lr){1-2}
For every divisor $d$, compute
$\mathrm{cnt}[d] = |\{j : d \mid A_j\}|$. &
Multiples-sieve divisor counts \\
\cmidrule[0.25pt](lr){1-2}
For each $A_i$, output the largest divisor $d \mid A_i$ with
$\mathrm{cnt}[d] \ge K$. &
Per-element feasible divisor search \\
\bottomrule
\end{tabularx}
\vspace{-8pt}
\end{wraptable}

\paragraph{Synthetic problem generation.}
The proposer generates each synthetic problem in two stages. First, \emph{solution proposal} conditions on $x^\star$ and the recent synthetic problem history $\mathcal{H}$, then produces a draft synthetic problem and a Python reference program. Second, \emph{problem refinement} reconstructs a clean problem statement, input/output specification, example I/O, and held-out input tests from the reference program. The held-out outputs are obtained by executing the reference program. This ordering makes the tests and statement conform to the program's actual behavior, rather than requiring a single decoding pass to keep the statement, tests, and solution mutually consistent. The subgoal generation prompt templates are listed in Appendix \ref{app:prompts}.

We then apply \textsc{Filter}, which rejects candidates with timeouts, runtime errors, duplicate-only test sets, or constant-output behavior. Accepted problems enter the fresh-problem set $\mathcal{D}_t$ and are added to the history after the solver attempts them. Invalid candidates do not enter the solver batch and receive a fixed proposer penalty.

\paragraph{Synthetic problem history.}
TTSP maintains a history $\mathcal{H}$ of accepted synthetic problems and the solver's most recent rollouts on them. At each iteration, the solver batch mixes new synthetic problems $\mathcal{D}_t$ with a buffer of previous problems $\mathcal{B}_t$ sampled from $\mathcal{H}$. The sampler favors problems whose current solve rate lies in a learnability window: already-solved problems provide little training signal, while near-impossible problems produce nearly all-zero rewards. Since \textsc{UpdateHistory} overwrites each problem's latest rollouts and solve rate, both proposer contexts and buffer sampling track the current solver rather than a stale snapshot.

\paragraph{Rewards.} The solver receives a binary synthetic reward
\[
r_\theta(\hat{y}, \hat{x}) = v_{\mathrm{syn}}(\hat{y}; \hat{x}),
\]
which equals one if and only if the solver rollout $\hat{y}$ passes all held-out generated tests for synthetic problem $\hat{x}$.

The proposer is rewarded for generating problems near the solver's current frontier. Let $\bar{r}_\theta(\hat{x})$ be the mean solver reward over rollouts on $\hat{x}$. For accepted synthetic problems, we define
\begin{equation}
r_\phi(\hat{x}) =
1 - 2\left|\bar{r}_\theta(\hat{x}) - \tfrac{1}{2}\right|,
\label{eq:proposer-reward}
\end{equation}
with a fixed penalty for invalid generations. The triangular profile targets the region where sparse Bernoulli rewards induce the largest group-relative signal under GRPO, so the proposer policy is rewarded for keeping the solver policy near its frontier \citep{hubotter2025learningonjob}. The buffer eligibility window is chosen to match the high-reward part of this profile.

\paragraph{Optimization.}
We optimize both adapters with GRPO \citep{shao2024grpo}, DAPO-style asymmetric clipping \citep{yu2025dapo}, and zero KL regularization following the test-time adaptation setup of \citet{hubotter2025learningonjob}. Hyperparameters are listed in Appendix \ref{app:hyperparams}.

\section{Experiments}
\label{sec:experiments}

\subsection{Experimental Setup}
\label{sec:setup}

\begin{wraptable}{r}{0.5\textwidth}
\vspace{-12pt}
\centering
\small
\setlength{\tabcolsep}{3pt}
\renewcommand{\arraystretch}{1.2}
\caption{\textbf{Required verifier access per method.}
Each method requires test case inputs, test case outputs, neither, or both, from which a learning signal can be derived.}
\label{tab:method-access}
\begin{tabularx}{\linewidth}{@{}X>{\centering\arraybackslash}X@{}}
\toprule
\textbf{Method} & \textbf{Verifier Access} \\
\midrule
Rejection sampling   & Inputs + Outputs \\
Context optimization & Inputs + Outputs \\
Majority voting      & Inputs Only \\
\midrule
TTSP (ours)          & None \\
\bottomrule
\end{tabularx}
\vspace{-19pt}
\end{wraptable}

\paragraph{Task domain.}
We evaluate TTSP in the domain of competitive programming. We use the benchmark LiveCodeBench v6 \citep{jain2024livecodebench}, which contains 131 target problems released between February and May 2025, after the Qwen3-8B knowledge cutoff. The benchmark provides public test cases, private test cases, and a broad range of problem difficulties. This lets us vary verifier strength and target problem difficulty within a single controlled evaluation setting.

\paragraph{Target problem set.}
From LiveCodeBench v6, we keep problems satisfying $0<\mathrm{mean}@512<0.7$ for Qwen3-8B and name the resulting target problem set HardLCBv6. Qwen3-8B has a wide range of non-zero accuracies over this set of problems.

We further split HardLCBv6 into MidHardLCBv6, containing target problems with $0.1<\mathrm{mean}@512<0.7$, and VeryHardLCBv6, containing target problems with $0 < \mathrm{mean}@512 \le 0.1$. We evaluate TTSP and baselines on these subsets of LiveCodeBench v6. Qwen3-8B solve rates for each target problem are listed in Appendix~\ref{app:dataset}.

% Figure~\ref{fig:regime-diagram} summarizes this empirical partition of target problems by base model coverage.

% \begin{figure}[t]
% \centering
% \includegraphics[width=0.68\textwidth]{main_figures/regime_diagram.png}
% \caption{\textbf{Regimes for per-target test-time scaling.}
% We group target problems by the empirical coverage of correct answers under the base policy, measured by mean@512. Our experiments evaluate targets with nonzero but imperfect base coverage, $0 < \mathrm{mean}@512 < 0.7$. Within this range, we separate a near-zero-coverage subset from the consistency regime, where correct rollouts occur but are not produced reliably. TTSP's gains are concentrated in the consistency regime, while the near-zero-coverage subset is reported separately and shows little or no improvement. The cutoffs at $0.1$ and $0.7$ are empirical thresholds used to define our LiveCodeBench evaluation subsets.}
% \label{fig:regime-diagram}
% \end{figure}

\begin{figure}[t]
\centering
\includegraphics[width=\textwidth]{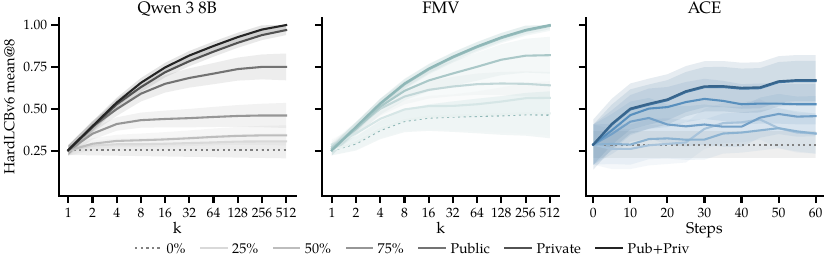}
\caption{\textbf{Baselines performance on HardLCBv6 as compute scales.} Each line reports the mean@8 with the indicated evaluation-time verifier applied to the validation problem submissions. Left and middle: Qwen 3 8B and FMV scale on the number of sampled rollouts k. Right: ACE scales on the number of generation–reflection–curation steps.}

\label{fig:baseline-scaling}
\end{figure}

\paragraph{Verifier construction.}
We evaluate seven verifier conditions. Four are artificial verifiers, $v_0$, $v_{25}$, $v_{50}$, and $v_{75}$, calibrated to reject the indicated fraction of incorrect sampled solutions; see Appendix~\ref{app:artificial-verifier-calibration}. Three are LiveCodeBench verifiers: public test cases ($v_{\mathrm{pub}}$), private test cases ($v_{\mathrm{priv}}$), and public + private test cases ($v^\star$). The verifier $v_0$ accepts every rollout, while $v^\star$ is the oracle verfier. These enable a closer investigation of a test-time scaling method's performance across various levels of weak (and absent, in the case of $v_0$) to strong (in the limit, $v^\star$) verifiers.

\subsection{Evaluation Protocol}
\label{sec:evaluation-protocol}

Each method output is evaluated as a set of eight target problem submissions. 
For aggregate comparisons, this set comes from the converged or final method output; for training-curve diagnostics, the same definition is applied at each checkpoint $t$. 
Let $Y=\{y_1,\ldots,y_8\}$ denote the eight submitted candidates for $x^\star$. 
For a verifier $v \in \{v_0, v_{25}, v_{50}, v_{75}, v_{\mathrm{pub}}, v_{\mathrm{priv}}, v^\star\}$, we form the admitted subset
\[
A(v)=\{y\in Y : v(y)=1\}.
\]
For non-empty admitted sets, we report the filtered mean@8 score
\[
\mathrm{mean@}8(v)
=
\frac{1}{|A(v)|}
\sum_{y\in A(v)} v^\star(y).
\]
Under $v_0$, no candidate is filtered out, so $A=Y$ and mean@8 is the fraction of all eight candidates that pass $v^\star$. Throughout the paper, mean@8 denotes this filtered score; we write unfiltered mean@8 for the fraction of all eight submitted candidates that pass $v^\star$ with no evaluation-time filtering. Empty admitted sets are handled by the carry-forward rule in Appendix~\ref{app:filtered-evaluation}. Since the same filtering pass is applied to all methods, differences in mean@8 reflect differences in candidate quality rather than differences in scoring.

\subsection{Baselines}
\label{sec:baselines}

We compare TTSP against one representative method from each test-time scaling family in Table~\ref{tab:method-access}: rejection sampling, majority voting, and context optimization. 
The table records the verifier access required by each method. The common filtered mean@8 protocol from Section~\ref{sec:evaluation-protocol} is then applied uniformly to all methods.

\paragraph{Rejection sampling.}
Our direct filtering baseline samples candidates from the unadapted Qwen3-8B policy. The model produces eight submissions for $x^\star$, and the verifier is used only in the evaluation-time filtering pass described above. This baseline measures how much performance can be obtained from base-policy sampling without any prompt or weight adaptation.

% \paragraph{Base model rejection sampling (Qwen 3 8B).} The unadapted base model produces the $8$ candidates on $x^\star$. No reward signal is used during the run, and the verifier enters only through the evaluation-time filtering pass above. For this baseline, filtering on the base policy's candidates with no preceding adaptation is the entire method, which is the standard rejection sampling baseline in test-time scaling.

% \paragraph{Functional Majority Voting (FMV) \citep{launer2026fmv}.} We run FMV on rollouts from the base policy. For each verifier-strength condition, FMV first filters candidate submissions using the corresponding verifier. It then clusters the verifier-admitted candidates by their execution behavior on the remaining public+private test inputs and selects the modal functional cluster. FMV therefore uses verifier outcomes for pre-selection and test inputs for consensus, but performs no weight or prompt update. The evaluation-time filtering pass is then applied to FMV's submissions.

\begin{figure}[t]
\centering
\begin{minipage}[t]{0.485\textwidth}
\centering
\includegraphics[width=\linewidth]{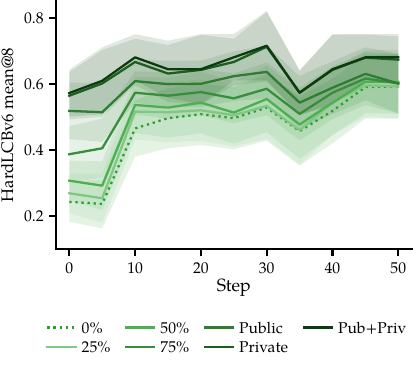}
\caption{\textbf{Performance of TTSP on HardLCBv6 over training steps.} Each line reports the mean@8 at different evaluation-time verifier strengths applied to the solver policy's target problem submissions. Shaded bands are 90\% bootstrap confidence intervals over target problems.}
\label{fig:ttt_results_aggregate_hard}
\end{minipage}\hfill
\begin{minipage}[t]{0.485\textwidth}
\centering
\includegraphics[width=\linewidth]{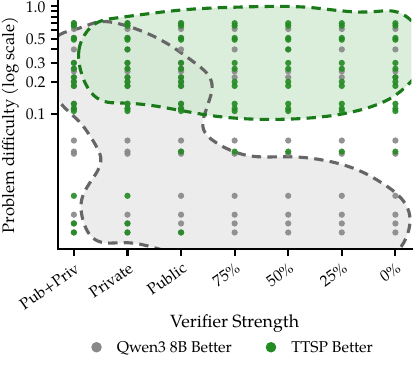}
% \caption{\textbf{Best performing method as a function of target problem difficulty and verifier strength.} Each point is a (target problem, verifier strength) pair, colored by whichever method yields higher mean@8. TTSP dominates the upper region (non-trivial base coverage, weak verifier), whereas base model with rejection sampling dominates the bottom strip (very low coverage) and left edge (strong verifier).}
\caption{\textbf{Best performing method per problem and verifier strength.}
Each point is a (target problem, verifier strength) pair, colored by the method with higher mean@8. The x-axis orders verifier strength and the y-axis reports problem difficulty, measured as the base policy's mean@512.}
\label{fig:phase_diagram}
\end{minipage}
\end{figure}

\paragraph{Majority voting.}
Our consensus baseline is Functional Majority Voting (FMV) \citep{launer2026fmv}. 
The FMV voting rule itself is input-only: it clusters candidate programs by execution behavior on target problem input cases and selects the modal functional cluster. 
In the verifier-strength sweep, we combine this input-only voting rule with the current verifier condition by first discarding candidates that fail the verifier and then applying FMV to the admitted pool. 
This baseline measures whether target problem rollouts contain a recoverable correct functional cluster after the available verifier has removed some incorrect candidates. 
Appendix~\ref{app:fmv-baseline} gives the exact filtering, scoring, and tie-handling procedure.

TTRL \citep{zuo2025ttrl}, while well-known for its majority voting reward, is designed for benchmark-level adaptation, jointly training on a population of test problems via per-prompt majority vote rewards, unlike our per-problem setting. TTRL is also only evaluated on math, whereas FMV was created for competitive programming, in which majority voting can be done in more strategic ways due to the existence of test cases.

\paragraph{Context optimization.}
Our prompt-optimization baseline is ACE \citep{zhang2026ace}. We run ACE separately at each verifier strength, using the pass/fail outcome under that verifier as the reward signal for its generator--reflector--curator loop. The evaluation-time filtering pass is then applied to ACE's final submissions. We use ACE rather than GEPA \citep{gepa2025} because ACE was the stronger context-optimization variant in our preliminary runs. Full implementation details are provided in Appendix~\ref{app:ace-baseline}.

\subsection{Results}
\label{sec:main-results}

\paragraph{Robustness across verifier strengths.} TTSP is the best-performing method at every verifier strength we evaluate (see Figure \ref{fig:verifier_strength_hard}). The TTSP curve is approximately flat across the full sweep, while all three baselines decline as the verifier weakens and the gap to TTSP widens accordingly. TTSP’s training-time reward is intrinsic to the self-play loop, so the verifier strength axis varies only at evaluation time; the baselines all depend on the verifier either for selection (rejection sampling, FMV) or as a reward signal (ACE), and their performance is bounded by the verifier’s discriminative power.

\paragraph{TTSP training dynamics.}
Figure~\ref{fig:ttt_results_aggregate_hard} shows TTSP performance over training steps. Most of the improvement occurs within the first $10$ training steps, after which the curves remain stable up to checkpoint-to-checkpoint noise. The spread between verifier-strength conditions also narrows during training. These dynamics suggest that the synthetic curricula quickly shift probability mass toward candidates that remain useful under later evaluation-time filtering.

\paragraph{Scaling baseline methods.}
Figure~\ref{fig:baseline-scaling} reports a scaling study along each baseline's native compute axis. The Qwen3-8B rejection sampling and FMV panels vary the rollout budget $k$, while the ACE panel varies the number of context optimization iterations and scores the current ACE submission. Qwen3-8B rejection sampling improves with larger rollout budgets primarily when the verifier is strong enough to reject incorrect candidates. FMV improves with additional rollouts because functional agreement becomes more informative. ACE improves over iterations only when the verifier provides informative pass/fail feedback; under $v_0$, it remains at the base-policy level because no optimization steps are taken. These results demonstrate that increasing compute does not remove the baselines' dependence on target-verifier information.

\paragraph{Verifier-free setting.}
In the $v_0$ setting, the verifier admits every candidate. In this setting, ACE is unable to improve because it is not provided any verification feedback. FMV remains limited to selecting among target problem rollouts sampled from the base policy, while TTSP improves substantially over all baselines.

\begin{wrapfigure}{r}{0.5\textwidth}
\vspace{-12pt}
\centering
\includegraphics[width=\linewidth]{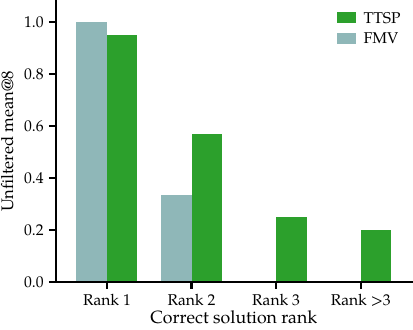}
% \caption{\textbf{Unfiltered mean@8 of TTSP and FMV binned by rank of the correct functional cluster in the base policy.} FMV performs strongly for rank 1 problems, drops sharply at rank 2, and is unable to identify the correct solution for higher ranks. TTSP retains non-trivial accuracy across all rank bins.}
\caption{\textbf{Unfiltered mean@8 of TTSP and FMV binned by rank of the correct functional cluster in the base policy.}
Bars group target problems by the rank of the correct functional cluster among base-policy rollouts.}
\label{fig:fmv_ranks}
\vspace{-20pt}
\end{wrapfigure}

\paragraph{Majority voting and solution rank.} % \paragraph{Consensus selection and modal rank.}
Figure~\ref{fig:fmv_ranks} bins target problems by the rank of the correct functional cluster in the base policy's rollout distribution. FMV is strongest when the correct cluster is modal, slightly exceeding TTSP performance. When the correct cluster is ranked second, FMV drops sharply and TTSP overtakes it. When the correct cluster has rank at least three, FMV is unable to identify the correct solution while TTSP retains non-trivial accuracy. We can observe that, while TTSP is impacted by correct solution output rank, it degrades gracefully as rank increases, whereas FMV is fully reliant on the correct solution being approximately the modal output.

\paragraph{TTSP is limited by target problem difficulty.}

TTSP requires the correct rollout to exist in the base policy’s support with non-trivial mass. Splitting HardLCBv6 into MidHardLCBv6 and VeryHardLCBv6 reveals a sharp cutoff (Figure \ref{fig:ttt_delta}). On MidHardLCBv6, the improvement of TTSP over the base policy is small at strong verifiers and rises sharply as the verifier weakens. On VeryHardLCBv6, the improvement is approximately zero across the entire sweep — slightly negative at the strongest verifiers, where rejection sampling on the base policy can still surface the rare correct rollout, and only marginally positive at the weakest. Below $\mathrm{mean}@512 \approx 0.1$, the proposer policy cannot bootstrap a curriculum that composes back to $x^\star$.

\paragraph{When is TTSP most effective?}
Figure~\ref{fig:phase_diagram} summarizes the same pattern at the per-target level. TTSP dominates when base model coverage is non-trivial and the evaluation-time verifier is weak. Base model rejection sampling dominates when the verifier is strong enough to identify rare correct rollouts or when base model coverage is too low for TTSP to build a useful curriculum. The resulting boundary places TTSP in the consistency regime: it is most useful when correct or partially useful target-related behavior is present but not reliable under direct sampling.

\section{Discussion}
\label{sec:discussion}

Test-Time Self-Play (TTSP) shifts the source of learning signal for test-time scaling. Existing methods derive their selection or learning signal from a verifier on the target problem, and their effectiveness degrades as that verifier weakens. In TTSP, a solver policy is instead trained on execution feedback from synthetic problems generated by a proposer policy (both initialized from the same base policy), with the target problem verifier left out of the training loop. The result is a test-time scaling algorithm that is substantially more robust to weak and absent target verifiers, conditional on the base policy having non-trivial coverage of the target problem $x^\star$.

\begin{wrapfigure}{r}{0.5\textwidth}
\vspace{-12pt}
\centering
\includegraphics[width=\linewidth]{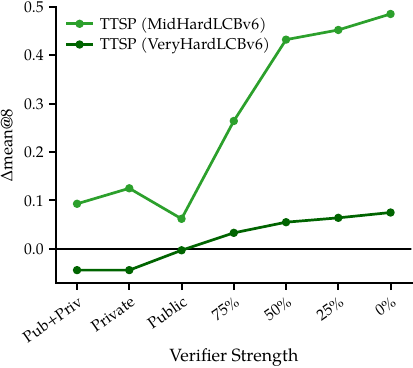}
% \caption{\textbf{Difference in mean@8 between TTSP and Qwen3-8B on HardLCBv6 vs. VeryHardLCBv6.} The gap is large and positive on HardLCBv6 as the verifier weakens, whereas TTSP only slightly outperforms the base policy on VeryHardLCBv6.}
\caption{\textbf{Difference in mean@8 between TTSP and Qwen3-8B.}
The gap is large and positive on MidHardLCBv6 as the verifier weakens. On VeryHardLCBv6, the gap is near zero: only marginally positive under the weakest verifiers.}
\label{fig:ttt_delta}
\vspace{-14pt}
\end{wrapfigure}

\paragraph{Limitations.} TTSP adapts per target problem and does not amortize compute across targets, suiting deployment contexts where each target is valuable enough to merit a dedicated adaptation budget. TTSP also only addresses the consistency regime: when the base policy has effectively zero coverage of the target answer, no curriculum the proposer policy can synthesize will compose back to it (Figure~\ref{fig:ttt_delta}). Furthermore, like other self-play algorithms, TTSP derives its training signal entirely from self-generated content rather than from any externally validated target. This makes training-time stability a concern that should be actively monitored. The proposer policy can drift toward synthetic problems that hack its reward function rather than ones that compose back to $x^\star$, and the solver policy's entropy can collapse under sustained training on a narrow distribution \citep{bailey2026sgs}. See Appendix \ref{app:dynamics} for further discussion of training dynamics.

\paragraph{Future work.} Several natural extensions follow from relaxing TTSP's constraints. The proposer policy's synthetic problem set can be augmented with $x^\star$ itself whenever a non-degenerate target problem reward exists, combining curriculum-driven and target-driven training in one system. The set can also be augmented with nearest-neighbor synthetic problems retrieved from external corpora when those are available \citep{hubotter2025learningonjob}, ingesting auxiliary data without losing the curriculum structure. Furthermore, TTSP could be applied to other domains with rich tools, including domains with execution sandboxes, simulation engines, or database engines. Finally, tackling the discovery regime, where the base model has near-zero coverage of the target problem solution, remains an important open direction; it likely requires either external information or substantially stronger search than the curriculum alone provides.

\ifcolmfinal
\section*{Acknowledgements}
We thank Leander Diaz-Bone, Antonio Mari, Anton Baumann and Catherine Hou for helpful discussions. AY was supported by Tufa Labs AG. JH was supported by the Swiss National Science Foundation under NCCR Automation, grant agreement 51NF40 180545. IH was supported by an ETH AI Center Postdoctoral Fellowship.

This project was supported through the Swiss AI compute grant a156.
\else\ifcolmpreprint
\section*{Acknowledgements}
We thank Leander Diaz-Bone and Antonio Mari for helpful discussions and Catherine Hou for stylistic feedback. AY was supported by Tufa Labs AG. JH was supported by the Swiss National Science Foundation under NCCR Automation, grant agreement 51NF40 180545. IH was supported by an ETH AI Center Postdoctoral Fellowship.

This project was supported through the Swiss AI compute grant a156.
\fi\fi

\bibliography{references}
\bibliographystyle{colm2026_conference}

%%%%%%%%%%%%%%%%%%%%%%%%%%%%%%%%%%%%%%%%%%%%%%%%%%%%%%%%%%%%%%%%
% APPENDIX
%%%%%%%%%%%%%%%%%%%%%%%%%%%%%%%%%%%%%%%%%%%%%%%%%%%%%%%%%%%%%%%%

\captionsetup[figure]{skip=10pt}% restore default for the appendix
\mainbodydisplaysfalse\abovedisplayskip=\savedADS \belowdisplayskip=\savedBDS \abovedisplayshortskip=\savedADSS \belowdisplayshortskip=\savedBDSS % appendix keeps the template's display spacing
\setlength{\textfloatsep}{20pt plus 2pt minus 4pt}% restore template default for the appendix

\clearpage
\appendix
% Appendix: figure-only pages start at the top of the page instead of being vertically centred
\makeatletter\setlength{\@fptop}{0pt}\setlength{\@fpsep}{12pt plus 2fil}\setlength{\@fpbot}{0pt plus 1fil}\makeatother

\section{Target Problems and Evaluation Protocol}
\label{app:eval-protocol}

This appendix fixes the evaluation substrate used throughout Section~\ref{sec:experiments}. We first list the target problem subsets, then define the verifier-strength axis and the common filtered evaluation metric used to compare all methods.

\subsection{Target problem subsets}
\label{app:dataset}

Section~\ref{sec:setup} defines the evaluation set by filtering a recent slice of LiveCodeBench v6 to problems where Qwen 3 8B has non-zero but imperfect base coverage \citep{jain2024livecodebench,yang2025qwen3}. This appendix lists the exact target problem identifiers and base model mean@512 values. The split is used throughout the paper: MidHardLCBv6 is the consistency-regime subset where TTSP can exploit non-trivial base coverage, and VeryHardLCBv6 is the near-discovery subset where base coverage is too small for the synthetic curriculum to compose reliably back to $x^\star$.

\begin{table}[!htbp]
\centering
\caption{\textbf{MidHardLCBv6 target problems.} LiveCodeBench v6 target problem identifiers and base model mean@512 of Qwen 3 8B for each selected target problem in the MidHardLCBv6 subset ($0.1 < \mathrm{mean}@512 < 0.7$).}
\begin{tabular}{rrr}
\toprule
Target Problem & mean@512 & Correct Solution Rank \\
\midrule
1035 & 0.686 & 1 \\
1050 & 0.645 & 1 \\
976 & 0.633 & 1 \\
920 & 0.596 & 1 \\
1047 & 0.555 & 1 \\
962 & 0.523 & 1 \\
916 & 0.500 & 1 \\
958 & 0.496 & 1 \\
948 & 0.488 & 1 \\
918 & 0.426 & 1 \\
926 & 0.297 & 2 \\
934 & 0.250 & 2 \\
933 & 0.219 & 2 \\
921 & 0.201 & 2 \\
980 & 0.188 & 2 \\
957 & 0.186 & 2 \\
987 & 0.135 & 3 \\
909 & 0.125 & 2 \\
960 & 0.111 & 4 \\
\bottomrule
\end{tabular}
\end{table}

\begin{table}[!htbp]
\centering
\caption{\textbf{VeryHardLCBv6 target problems.} LiveCodeBench v6 target problem identifiers and base model mean@512 of Qwen 3 8B for each selected target problem in the VeryHardLCBv6 subset ($0 < \mathrm{mean}@512 \le 0.1$).}
\begin{tabular}{rrr}
\toprule
Target Problem & mean@512 & Correct Solution Rank \\
\midrule
941 & 0.057 & 3 \\
1017 & 0.045 & 2 \\
946 & 0.043 & 3 \\
919 & 0.018 & 3 \\
1027 & 0.012 & 2 \\
947 & 0.010 & 11 \\
905 & 0.010 & 4 \\
979 & 0.004 & 37 \\
963 & 0.002 & 21 \\
\bottomrule
\end{tabular}
\end{table}

The remaining subsections make the verifier-strength axis explicit. The distinction is important for interpreting Figure~\ref{fig:verifier_strength_hard}: TTSP receives training rewards from execution on proposer-generated synthetic problems, whereas the baselines select, cluster, or optimize directly against information from the target problem verifier.

\subsection{Verifier-strength axis}
\label{app:verifier-axis}

A verifier is a selection rule applied to candidate solutions for the original target problem $x^\star$. Strong verifiers reject a large fraction of incorrect candidate programs, while weak verifiers admit many false positives. We evaluate across a sweep of verifier strengths to factor the test-time scaling problem into candidate generation and candidate selection, following the main-text motivation that test-time compute alone is not enough when the selection signal is weak \citep{snell2024scaling,brown2024monkeys}.

The weakest condition is $v_0$, defined by $v_0(y)=1$ for every rollout $y$. It corresponds to no filtering: all generated candidates are treated as admissible, and the method is evaluated solely by how often its unfiltered candidates pass the oracle $v^\star$. The calibrated artificial verifiers $v_{25}$, $v_{50}$, and $v_{75}$ are constructed to reject approximately the indicated fraction of incorrect calibration rollouts. The natural LiveCodeBench verifiers $v_{\mathrm{pub}}$, $v_{\mathrm{priv}}$, and $v^\star$ are built from public tests, private tests, and their union, respectively \citep{jain2024livecodebench}. The public+private verifier $v^\star$ is both the strongest evaluation-time filter and the oracle used to measure final correctness.

\subsection{Natural LiveCodeBench verifiers}
\label{app:natural-verifiers}

LiveCodeBench provides public and private test cases for each target problem \citep{jain2024livecodebench}. We use these test suites as natural verifier-access levels. The public verifier $v_{\mathrm{pub}}$ models the realistic programming-contest or live-coding setting in which only sample or public tests are available at deployment. The private verifier $v_{\mathrm{priv}}$ and the public+private verifier $v^\star$ provide stronger selection rules and serve as high-verifier-access reference points.

All reported correctness scores are computed against $v^\star$. Thus, even when the evaluation-time filter is weaker than $v^\star$, the final measurement asks whether an admitted candidate passes the full public+private evaluator. This convention keeps the target quantity fixed across verifier-strength conditions and makes the curves in Figure~\ref{fig:verifier_strength_hard} comparable.

\subsection{Artificial verifier calibration}
\label{app:artificial-verifier-calibration}
\label{app:verifier-calibration}

The calibrated weak verifiers $v_{25}$, $v_{50}$, and $v_{75}$ are selected from a candidate pool containing retained generated tests as well as public and private LiveCodeBench tests. We call these verifiers artificial because the subset is chosen to match a target empirical rejection rate, not because every selected test is generated. Generated candidate inputs are produced using Qwen 3 8B \citep{yang2025qwen3} with the same refinement-style prompting pattern used by TTSP to reconstruct synthetic problem tests; the TTSP refinement prompt that this pattern mirrors is reported in Appendix~\ref{app:refinement_prompt}. These tests are not used to train TTSP. They are used only to construct controlled evaluation-time filters whose false-positive rate can be varied more finely than the natural public/private split.

Generated inputs do not come with trusted ground-truth outputs, so we label them by consensus among base model programs that are already correct under $v^\star$. Public and private LiveCodeBench tests use their provided labels. For each target problem, we sample $512$ Qwen 3 8B baseline rollouts and partition them into correct and incorrect programs using $v^\star$. For a generated input $z$, we execute the correct baseline programs on $z$. If the correct programs disagree, $z$ is discarded; if they agree, the consensus output is used as the reference label. This step is designed to reduce false negatives: a generated test is kept only when programs known to solve the target problem agree on its output.

We then reduce verifier construction to a coverage problem over the incorrect calibration rollouts. Each retained candidate test case $t_j$ has an associated rejection set
\[
E_j \subseteq \mathcal{Y}_{\mathrm{incorrect}},
\]
where $E_j$ contains the incorrect calibration rollouts that either fail to execute on $t_j$ or produce an output different from the consensus output. For a candidate verifier defined by a subset of tests $S$, its empirical strength is
\[
c(S)
=
\frac{
\left|\bigcup_{t_j \in S} E_j\right|
}{
|\mathcal{Y}_{\mathrm{incorrect}}|
}.
\]
This definition accounts for overlap between tests: two tests that reject the same incorrect rollouts do not increase verifier strength additively.

For each target problem, $v_{25}$, $v_{50}$, and $v_{75}$ are selected as subsets of the retained candidate tests whose empirical coverage $c(S)$ is closest to $0.25$, $0.50$, and $0.75$, respectively. The resulting filters intentionally permit false positives, because a weak verifier should accept some incorrect programs. We keep $v_{\mathrm{pub}}$, $v_{\mathrm{priv}}$, and $v^\star$ as separate natural verifier conditions rather than folding them into this calibrated artificial sweep.

\subsection{Filtered evaluation metric}
\label{app:filtered-evaluation}

Section~\ref{sec:evaluation-protocol} defines the filtered mean@8 metric. 
For iterative methods and training-curve diagnostics, the same metric is applied checkpoint-wise to the eight target problem submissions sampled at checkpoint $t$.

If the evaluation-time verifier admits no candidates for a given target problem, method, checkpoint, and verifier condition, the reported score is carried forward from the most recent earlier checkpoint with a non-empty admitted set for the same target problem, method, and verifier condition. 
If no earlier checkpoint has a non-empty admitted set, the score is defined to be $0$. 
For non-iterative baselines, this rule reduces to assigning score $0$ whenever the admitted set is empty. 
This convention avoids undefined scores when a strong verifier rejects all sampled candidates.

Table~\ref{tab:verifier-tests} reports how many tests are used by each evaluation-time verifier on HardLCBv6, averaged over the 28 target problems and grouped by test source.

\subsection{Uncertainty bands}
For Figures~\ref{fig:ttt_results_aggregate_hard} and~\ref{fig:baseline-scaling}, shaded bands denote 90\% nonparametric bootstrap confidence intervals over target problems. 
For each curve point, we resample the HardLCBv6 target problems with replacement $10{,}000$ times, recompute the mean score across the resampled target set, and plot the 5th and 95th percentiles of the resulting bootstrap distribution.

For the diagnostic curves in Figures~\ref{fig:dynamics_scores}, \ref{fig:dynamics_entropy}, and the right panel of Figure~\ref{fig:embedding_novelty_target_similarity}, shaded bands denote $\pm 1$ standard error of the mean across target problems, computed as the empirical standard deviation of target-level values divided by $\sqrt{n}$.

\begin{table}[!htbp]
\centering
\caption{\textbf{Tests used by each evaluation-time verifier on HardLCBv6.} Values are averages over the 28 target problems and are grouped by test source.}
\label{tab:verifier-tests}
\small
\setlength{\tabcolsep}{5pt}
\renewcommand{\arraystretch}{1.10}
\begin{tabular}{@{}lrrrr@{}}
\toprule
\textbf{Verifier} & \multicolumn{4}{c}{\textbf{Mean tests used}} \\ 
\cmidrule(l){2-5}
 & \textbf{Generated} & \textbf{Public} & \textbf{Private} & \textbf{Total} \\ 
\midrule
$v_0$ & 0.00 & 0.00 & 0.00 & 0.00 \\
$v_{25}$ & 0.36 & 0.07 & 0.86 & 1.29 \\
$v_{50}$ & 0.39 & 0.07 & 1.00 & 1.46 \\
$v_{75}$ & 0.57 & 0.18 & 0.79 & 1.54 \\
$v_{\mathrm{pub}}$ & 0.00 & 2.86 & 0.00 & 2.86 \\
$v_{\mathrm{priv}}$ & 0.00 & 0.00 & 37.50 & 37.50 \\
$v^\star$ & 0.00 & 2.86 & 37.50 & 40.36 \\
\bottomrule
\end{tabular}
\end{table}

% \subsection{Method-specific verifier access}
% \label{app:method-specific-verifier-access}

% All methods are scored with the same filtered evaluation protocol, but they use verifier information differently before that final filter is applied. The base model rejection-sampling baseline has no adaptation step: Qwen 3 8B samples candidates, and the verifier is used only as the post-hoc filter. FMV first applies the verifier to remove failing candidates, then clusters the verifier-admitted programs by their behavior on the remaining available test inputs and selects the modal functional cluster \citep{launer2026fmv}. ACE uses verifier pass/fail outcomes as the reward signal for its generator--reflector--curator context-optimization loop \citep{zhang2026ace}. TTSP uses no target problem verifier information during training; its solver rewards come only from execution against proposer-generated reference outputs on proposer-generated synthetic tests. The same filtered evaluation protocol is then applied to the candidate set produced by each method.

\clearpage
\section{TTSP Training Details and Hyperparameters}
\label{app:training-details}

Section~\ref{sec:method} describes TTSP at the algorithmic level: a proposer policy generates target-conditioned synthetic problems, a solver policy trains on them, and both policies receive execution-based rewards. This appendix records the implementation choices needed to reproduce the experiments. The design follows the main-text constraint that adaptation is performed separately for each target problem and that no target problem verifier is used during TTSP training.

Both roles are implemented as separate LoRA adapters over a frozen Qwen 3 8B backbone \citep{hu2022lora,yang2025qwen3}. The separation mirrors the asymmetric self-play setup in Algorithm~\ref{alg:ttsp-detailed}: the proposer optimizes for a useful curriculum, while the solver optimizes for pass/fail performance on generated synthetic problems. Both adapters are optimized with GRPO \citep{shao2024grpo}. For each prompt, the active policy samples a group of responses, assigns scalar rewards, computes group-normalized advantages, and performs a clipped on-policy update. We use asymmetric clipping with $(\epsilon_{\mathrm{low}}, \epsilon_{\mathrm{high}})=(0.20,0.28)$ following DAPO \citep{yu2025dapo}. We set the KL coefficient to zero following the test-time adaptation setup of \citet{hubotter2025learningonjob}, so the update is driven by target-specific rewards rather than by explicit regularization toward the initialization.

Each solver step uses a batch of $48$ synthetic problem slots: $16$ fresh problems generated by the current proposer and $32$ replayed problems sampled from the synthetic problem history. For each synthetic problem, the solver samples a group of $8$ rollouts and receives binary execution rewards on held-out generated tests. Replay eligibility is determined by the most recent recorded solver solve rate for each historical synthetic problem. When a replayed problem is sampled, stale rewards are not reused: the current solver policy generates fresh rollouts, pass/fail rewards are recomputed, and the stored solve rate used for future proposer context and replay eligibility is overwritten. The synthetic problem history has no fixed horizon; accepted synthetic problems remain eligible for replay whenever their most recent recorded solve rate lies in the learnability window.

\subsection{Hyperparameters}
\label{app:hyperparams}

Tables~\ref{tab:core_hyperparams} and~\ref{tab:curriculum_hyperparams} report the full TTSP training configuration. Unless otherwise stated, values are fixed across target problems. In particular, aggregate method-comparison figures use converged checkpoints rather than per-target hidden-test checkpoint selection, matching the evaluation protocol in Section~\ref{sec:evaluation-protocol}.

\begin{table}[!htbp]
\caption{\textbf{Core TTSP training hyperparameters.}
All values are fixed across target problems. The model is adapted per target through separate proposer and solver LoRA adapters, while the Qwen 3 8B backbone remains frozen.}
\label{tab:core_hyperparams}
\centering
\small
\setlength{\tabcolsep}{6pt}
\renewcommand{\arraystretch}{1.12}
\begin{tabular}{@{}p{0.38\textwidth}p{0.50\textwidth}@{}}
\toprule
\textbf{Setting} & \textbf{Value} \\
\midrule

\multicolumn{2}{@{}l}{\textit{Model}} \\
Backbone & Qwen3-8B \\
Trainable parameters & Separate proposer and solver LoRA adapters \\
LoRA rank / $\alpha$ & $32 / 32$ \\
LoRA target modules & All linear layers \\

\midrule
\multicolumn{2}{@{}l}{\textit{Optimization}} \\
RL objective & GRPO \\
Optimizer & AdamW \\
Learning rate & $1\times 10^{-4}$ \\
Warmup & $8$ steps \\
Clipping & $(\epsilon_{\mathrm{low}}, \epsilon_{\mathrm{high}})=(0.20,0.28)$ \\
KL coefficient & $0$ \\
Gradient clip norm & $1.0$ \\

\midrule
\multicolumn{2}{@{}l}{\textit{Training budget}} \\
Training length & $50$ GRPO steps per target problem \\
Checkpoint selection & Converged checkpoint; no best hidden-test checkpoint selection \\

\bottomrule
\end{tabular}
\end{table}

\begin{table}[!htbp]
\caption{\textbf{Curriculum, rollout, and replay hyperparameters.}
Fresh synthetic problems are generated by the proposer; replayed synthetic problems are sampled from history and re-evaluated under the current solver policy when replayed.}
\label{tab:curriculum_hyperparams}
\centering
\small
\setlength{\tabcolsep}{6pt}
\renewcommand{\arraystretch}{1.12}
\begin{tabular}{@{}p{0.38\textwidth}p{0.50\textwidth}@{}}
\toprule
\textbf{Setting} & \textbf{Value} \\
\midrule

\multicolumn{2}{@{}l}{\textit{Generation}} \\
Inference engine & vLLM \\
Sampling temperature & $1.0$ \\
Max prompt / response length & $8096 / 16384$ tokens \\
Proposer contexts per step & $2$ \\
Proposer rollouts per context & $8$ \\
Fresh synthetic problem slots per solver step & $16$ \\

\midrule
\multicolumn{2}{@{}l}{\textit{Solver batch}} \\
Synthetic problem batch size & $48$ per solver step \\
Batch composition & $16$ fresh + $32$ replayed synthetic problems \\
Solver rollouts per synthetic problem & $8$ \\
Solver reward & Binary pass/fail on $10$ held-out generated tests \\

\midrule
\multicolumn{2}{@{}l}{\textit{Proposer reward and replay}} \\
Proposer reward & Equation~\ref{eq:proposer-reward} \\
Invalid-generation penalty & $-0.5$ \\
Replay eligibility & Most recent recorded solver solve rate in $[0.1,0.8]$ \\
History size & No fixed cap \\
Replay refresh & Fresh solver rollouts, rewards, and recorded solve rate recomputed on replay \\

\midrule
\multicolumn{2}{@{}l}{\textit{Context and length limits}} \\
Proposer context window & Previous $5$ steps \\
Synthetic problems shown in context & $3$ \\
Synthetic problem max length & $1024$ tokens \\
Reference solution max length & $4096$ tokens \\

\bottomrule
\end{tabular}
\end{table}
\clearpage
\section{Baseline Implementation and Diagnostics}
\label{app:baseline-details}

This appendix records baseline-specific details that are separate from the shared filtered evaluation protocol in Appendix~\ref{app:eval-protocol}.

\subsection{ACE Baseline Implementation}
\label{app:ace-baseline}

We include ACE \citep{zhang2026ace} as a test-time context-optimization baseline. In contrast to TTSP, ACE does not update model weights. Instead, it maintains a problem-specific textual playbook, revises that playbook from execution feedback, and conditions later candidate generations on the revised playbook. ACE therefore tests whether gains can be obtained from context-only adaptation, rather than from TTSP's weight-updating policy adaptation.

For a controlled comparison, all ACE roles use the same base model as our TTSP experiments, \texttt{Qwen3-8B}. Each ACE iteration consists of several generation--reflection rounds. In each round, the Generator proposes a Python solution using the current playbook, the candidate is executed against the feedback tests available under the current verifier condition, and the Reflector turns the resulting execution signal into a concrete diagnosis of the attempt. After the rounds in an iteration are complete, the Curator converts the accumulated reflections into a structured update to the playbook, which is then used in the next iteration.

We evaluate ACE under the same feedback regimes as the other test-time baselines: public tests, private tests, public + private tests, and calibrated artificial verifier tests. The feedback exposed to ACE is restricted to the tests available in the corresponding condition. The full public + private evaluator $v^\star$ is used only for held-out monitoring after each round and iteration, and is not revealed to the playbook except in the public + private oracle condition.

ACE is evaluated in online mode with one evolving playbook per target problem. We use $15$ ACE iterations, at most $5$ generation--reflection rounds per iteration, deterministic decoding during playbook evolution, a maximum generation length of $16{,}384$ tokens, and a maximum model context length of $40{,}960$ tokens.

\subsection{Functional Majority Voting Baseline}
\label{app:fmv-baseline}

We implement Functional Majority Voting (FMV) following \citet{launer2026fmv}. 
The FMV selection rule is input-only: it uses target problem inputs to compare candidate programs by functional behavior, but does not require target problem outputs.

Let $\mathcal{C}=\{c_1,\ldots,c_N\}$ be candidate programs and let $\mathcal{X}=\{x_1,\ldots,x_K\}$ be the target problem input cases used for functional comparison. 
Executing candidate $c_i$ on input $x_k$ produces output string $o_{i,k}$. 
We discard candidates that fail to execute on the comparison inputs and denote the remaining valid candidates by $\mathcal{C}_{\mathrm{val}}$.

For each valid candidate $c_i \in \mathcal{C}_{\mathrm{val}}$, we compute the FMV score
\[
S(c_i)
=
\sum_{\substack{c_j \in \mathcal{C}_{\mathrm{val}}\\ j\neq i}}
\sum_{k=1}^{K}
\mathbf{1}\{o_{i,k}=o_{j,k}\}.
\]
This score counts pairwise functional agreement between candidate $c_i$ and the other valid candidates over the comparison inputs. 
The original FMV selection rule returns a candidate in
\[
\arg\max_{c_i\in\mathcal{C}_{\mathrm{val}}} S(c_i).
\]
In our implementation, when multiple candidates attain the maximum score, we keep all tied maximizers:
\[
\mathcal{C}_{\mathrm{FMV}}
=
\left\{
c_i \in \mathcal{C}_{\mathrm{val}}
:
S(c_i)=\max_{c_j\in\mathcal{C}_{\mathrm{val}}} S(c_j)
\right\}.
\]

For the verifier-strength comparison in Figure~\ref{fig:verifier_strength_hard}, we wrap this input-only FMV rule with the current verifier condition. 
Given verifier $v$, we first form the verifier-admitted candidate pool
\[
\mathcal{C}_v = \{c \in \mathcal{C}: v(c)=1\}.
\]
We then apply the FMV rule above to $\mathcal{C}_v$, using the available target problem input cases for functional comparison. 
This wrapper uses the same verifier condition as the rest of the evaluation sweep, while Table~\ref{tab:method-access} reports the access required by the FMV voting rule itself. 
The selected candidates are then passed to the common filtered evaluation protocol in Appendix~\ref{app:filtered-evaluation}; when the same verifier has already been used to form $\mathcal{C}_v$, this final pass is idempotent except for the shared empty-set handling convention.

\subsection{Functional Majority Voting Under Partial Test-Input Access}
\label{app:norejectionsampling}

Section~\ref{sec:main-results} compares TTSP, FMV, ACE, and rejection sampling under the same evaluation-time filtering protocol. We also report a diagnostic variant that removes the final post-hoc filter and varies only the target problem test inputs available to FMV for functional clustering. This separates FMV's input-access benefit from rejection-sampling effects.

In this diagnostic, every method is scored directly against $v^\star$. The access condition determines which test inputs FMV can use to form functional clusters, rather than which candidates are filtered after generation. TTSP does not use target problem test inputs during training, so its unfiltered score is independent of this access condition. FMV can improve when additional inputs separate incorrect functional clusters from the correct one, but it still cannot update the underlying candidate distribution.

\begin{figure}[!htbp]
    \centering
    \includegraphics[width=0.5\textwidth]{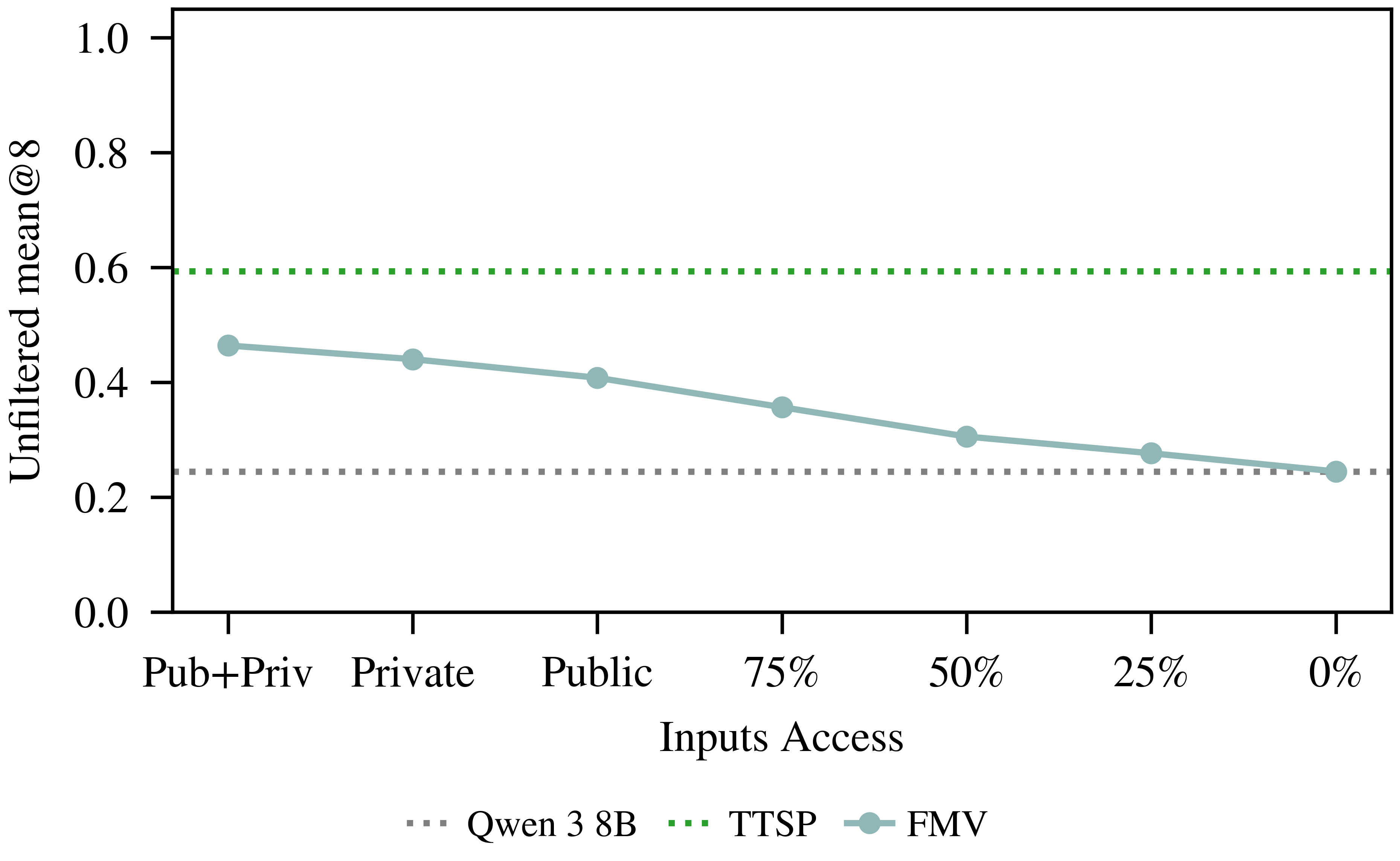}
        \caption{\textbf{FMV under partial test-input access on HardLCBv6.}
    The final evaluation-time filter is removed, and all submitted candidates are scored directly against the public+private evaluator $v^\star$. The x-axis denotes the test inputs available to FMV for functional clustering, not a post-hoc verifier. TTSP and Qwen 3 8B are shown as unfiltered baselines independent of FMV's input-access condition.}
    \label{fig:norejectionsamplingresults}
\end{figure}

\clearpage
\section{Synthetic Curriculum Quality and Dynamics}
\label{app:dynamics}

Self-generated training introduces two concrete stability risks. First, the proposer can generate invalid or uninformative synthetic problems. Second, the solver can over-specialize to a narrow generated distribution. This section reports diagnostics for those risks during TTSP adaptation on HardLCBv6. These diagnostics are logged only for analysis; they are not used as rewards, filters, or checkpoint-selection criteria.

\paragraph{Operational criteria.}
A synthetic problem is used for solver training only after it passes the validity filter in Section~\ref{sec:method}. Among accepted problems, the proposer reward in Equation~\ref{eq:proposer-reward} favors intermediate solver success, and the replay buffer samples problems whose most recent solver solve rate lies in the eligibility window in Appendix~\ref{app:training-details}. Thus the relevant checks are operational: rejected generations should be excluded before solver updates, proposer rewards should retain a non-degenerate high-reward tail, and the generated curriculum should not collapse to a single policy or program family. These checks address the stability concern noted in Section~\ref{sec:discussion}; similar concerns arise in recent data-free and self-play LLM training methods \citep{zhao2025absolutezero,huang2025rzero,chae2025coverage,bailey2026sgs}.

\subsection{Synthetic problem rejection diagnostics}
\label{app:rejection-diagnostics}

\begin{figure}[!htbp]
    \centering
    \includegraphics[width=\textwidth]{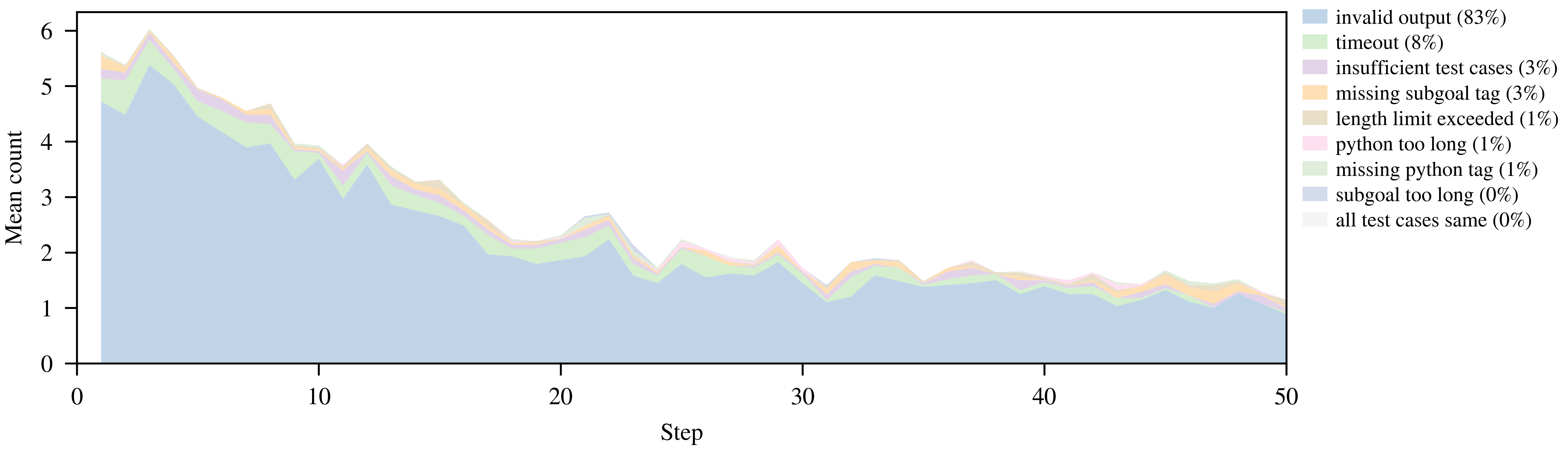}
    \caption{\textbf{Synthetic problem rejection diagnostics.}
    Pre-training proposer generations rejected before entering solver training, averaged over target problems and grouped by failure type. Rejection counts decrease over training; invalid generated output is the largest rejection category.}
    \label{fig:proposer_errors}
\end{figure}

Before a generated synthetic problem enters solver training, it must pass the validity filter described in Section~\ref{sec:method}. 
Figure~\ref{fig:proposer_errors} reports pre-solver-training rejections, averaged over target problems and decomposed by implementation-level failure type. 
Rejected generations are excluded from the solver batch and produce no solver rollout rewards; instead, they receive the fixed invalid-generation proposer penalty.

The failure types in Figure~\ref{fig:proposer_errors} refine the higher-level filter definition in Section~\ref{sec:method}. 
Invalid generated output and missing tags correspond to parsing failures; length-limit, \texttt{subgoal-too-long}, and \texttt{python-too-long} failures correspond to structural constraints; timeouts and runtime errors correspond to execution failures; and insufficient, duplicate, or all-identical test cases correspond to held-out-test degeneracy.

The total number of rejected generations decreases over training. 
Invalid generated output is the largest rejection category, followed by timeouts and insufficient test cases. 
This diagnostic shows that TTSP trains on a filtered subset of proposer generations rather than treating all generated problems as valid training data.

\subsection{Average and top-$k$ proposer reward}
\label{app:proposer-topk-reward}

The proposer update is based on sampled batches, so the batch mean alone can hide whether a step contains useful high-reward candidates. We therefore log both the mean proposer reward and the mean reward of the top four proposer samples at each step. A persistent gap between these curves indicates within-batch reward concentration: most samples may have low reward while a smaller subset still provides a non-degenerate proposer signal.

\begin{figure}[!htbp]
    \centering
    \includegraphics[width=0.5\textwidth]{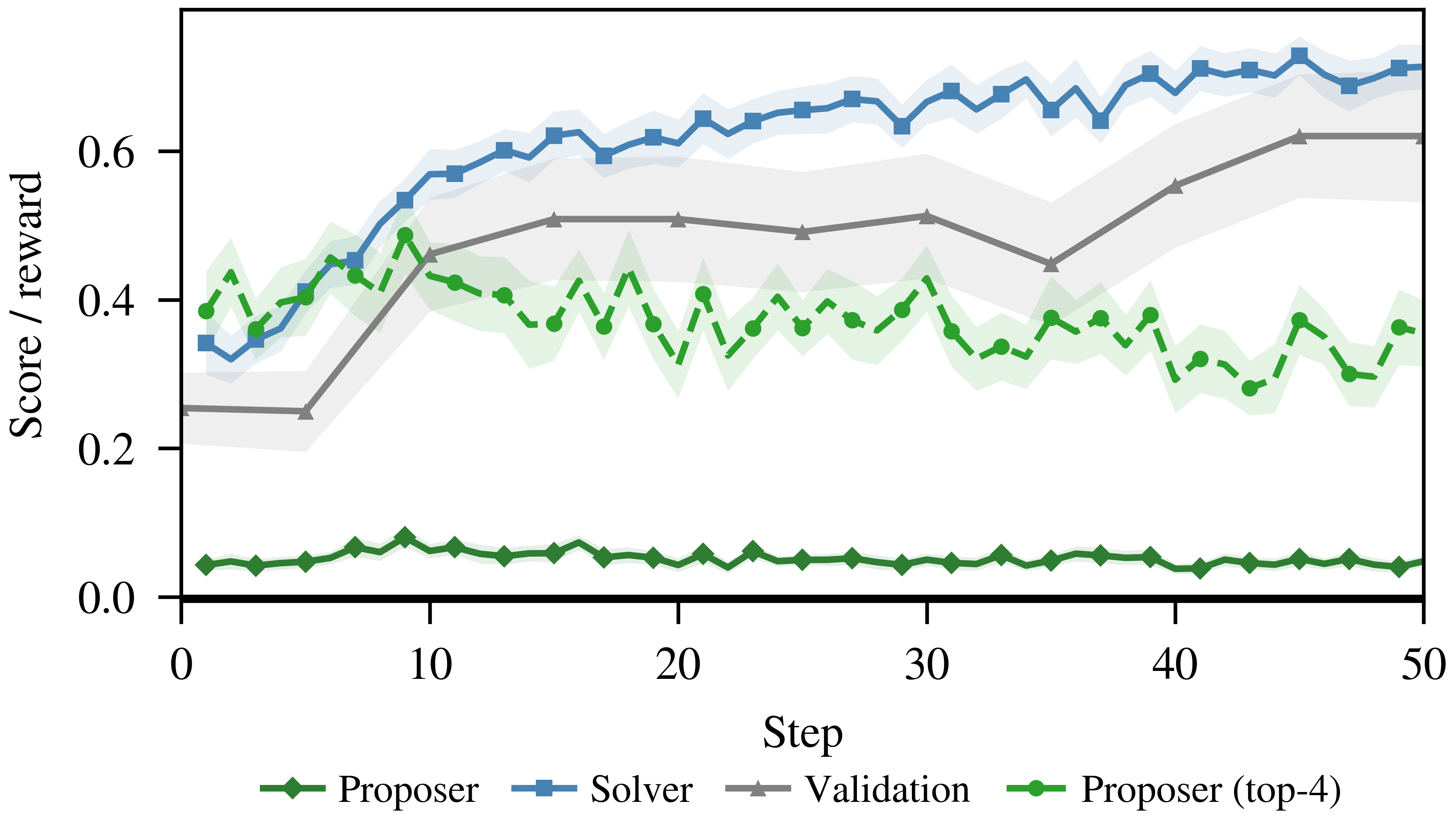}
    \caption{\textbf{Average and top-$4$ proposer reward during TTSP training.}
    The \emph{Proposer} curve is the mean proposer reward over generated synthetic problems, while \emph{Proposer (top-4)} averages the four highest proposer rewards at each step. 
    The gap measures within-batch reward concentration. Solver reward and unfiltered target validation are shown for reference. Shaded bands denote $\pm 1$ SEM across target problems.}
    \label{fig:dynamics_scores}
\end{figure}

Figure~\ref{fig:dynamics_scores} shows this pattern. The mean proposer reward remains low, while the top-$4$ reward is substantially higher. The solver reward also increases over training. Target validation denotes the solver's unfiltered mean@8 on the target problem under $v_0$. We use this plot as a non-degeneracy check for the synthetic reward signal; it does not imply that every generated synthetic problem is useful.

\subsection{Frozen-base and adapted-solver performance}
\label{app:frozen-base-synthetic}

To separate curriculum difficulty from solver adaptation, we evaluate both the frozen initialization and the current TTSP solver on the same synthetic problems generated at each step. The frozen model is Qwen 3 8B with no target-specific updates. Because the synthetic problem distribution changes over training, this is not a fixed-benchmark learning curve. Instead, it compares the adapted solver against the initialization on the current generated curriculum.

\begin{figure}[!htbp]
    \centering
    \includegraphics[width=0.5\textwidth]{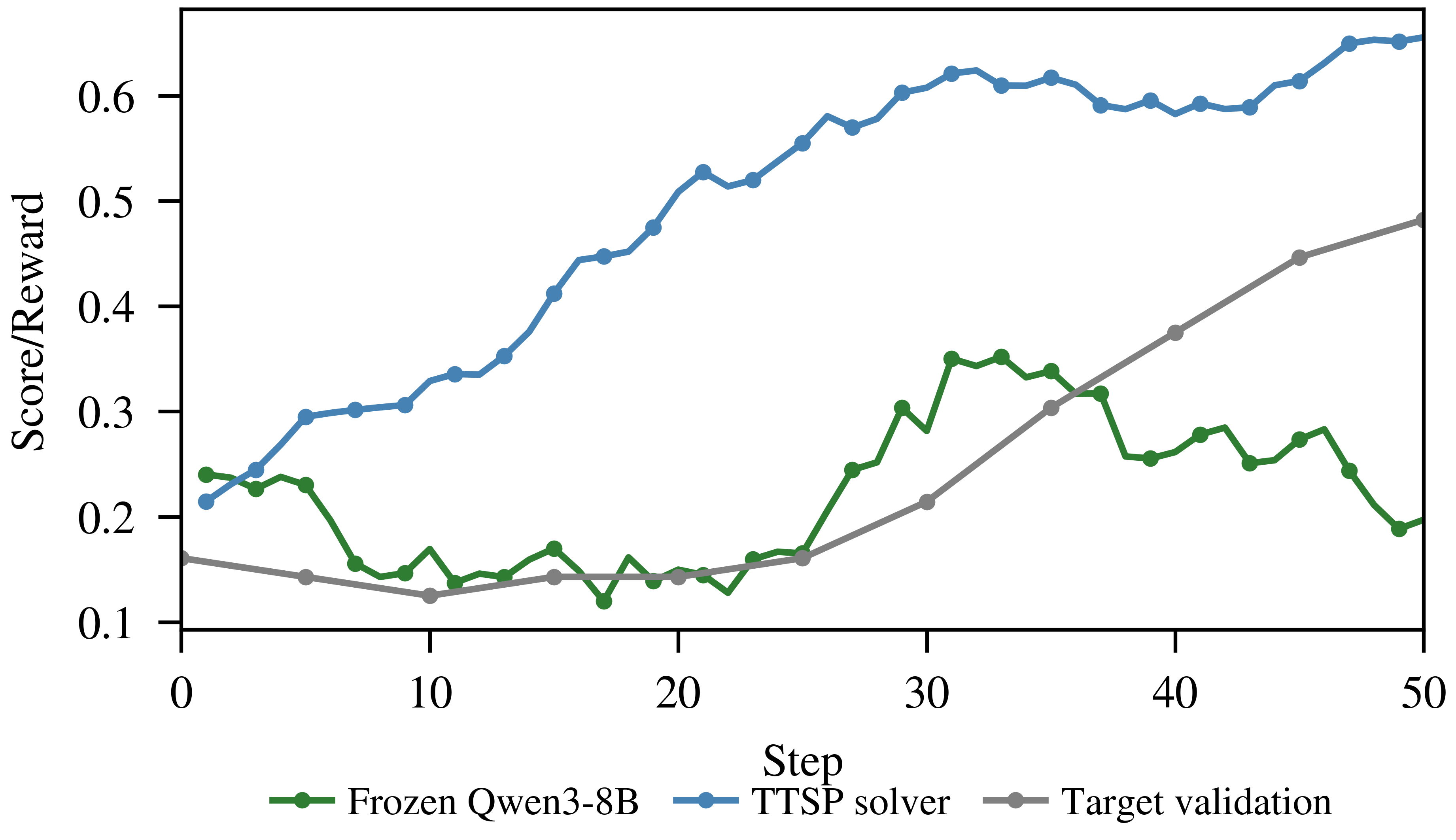}
    \caption{\textbf{Frozen-base and adapted-solver performance on generated synthetic problems.}
    Example trajectory for one target problem. \emph{Frozen Qwen 3 8B} is the initial model evaluated on the synthetic problems generated at each step; \emph{TTSP solver} is the adapted solver evaluated on the same synthetic problems; \emph{Target validation} is the adapted solver's unfiltered mean@8 on the original target problem under $v_0$.}
    \label{fig:dynamics_frozen_solver}
\end{figure}

Figure~\ref{fig:dynamics_frozen_solver} shows a representative trajectory. On the same generated synthetic problems, the adapted solver rises above the frozen initialization. This establishes an adaptation gain relative to the initial model on the generated curriculum. Target validation is plotted separately to show when this gain transfers back to the original target problem $x^\star$.

\subsection{Policy entropy}
\label{app:entropy-dynamics}

\begin{figure}[!htbp]
    \centering
    \includegraphics[width=0.5\textwidth]{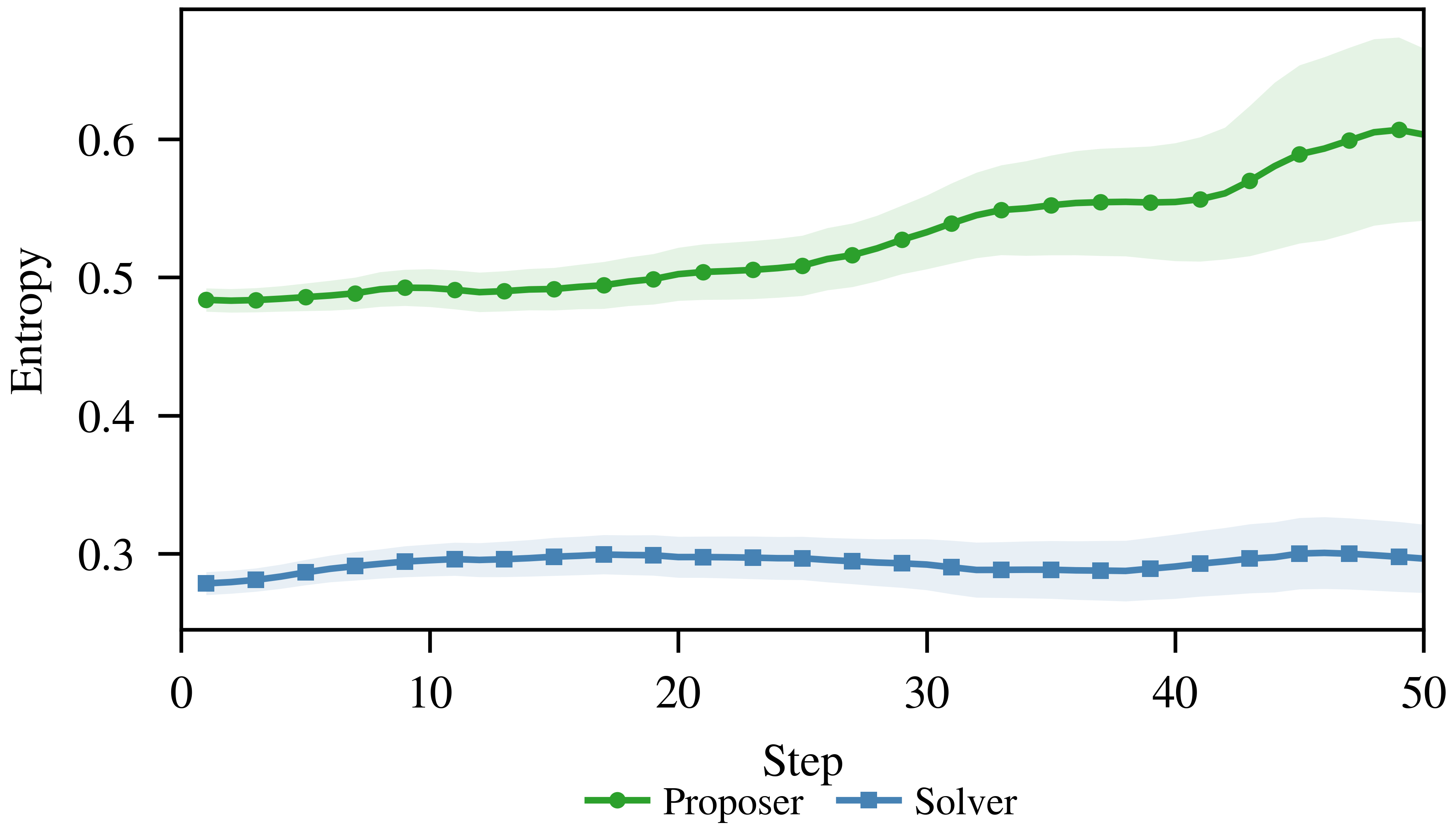}
    \caption{\textbf{Policy entropy during TTSP training.}
    Mean proposer-policy and solver-policy entropy across GRPO steps on HardLCBv6, averaged over target problems. The proposer entropy increases over training, while solver entropy remains stable. Shaded regions show variation across target problems. Shaded bands denote $\pm 1$ SEM across target problems.}
    \label{fig:dynamics_entropy}
\end{figure}

Figure~\ref{fig:dynamics_entropy} reports proposer and solver policy entropy across GRPO steps on HardLCBv6. Entropy is a distributional-collapse diagnostic. A sharp decrease in proposer entropy would indicate concentration of the synthetic problem generator, and a sharp decrease in solver entropy would indicate narrowing of the solver response distribution.

In our runs, proposer entropy increases gradually and solver entropy remains stable over the adaptation horizon. Entropy does not measure semantic novelty directly; the embedding diagnostics in Appendix~\ref{app:embedding-novelty-target-relatedness} audit program-level novelty separately.

\subsection{Embedding-based novelty and target relatedness}
\label{app:embedding-novelty-target-relatedness}

We audit accepted synthetic problems with two embedding-based diagnostics. First, for each valid synthetic problem, we embed its extracted Python reference program using \texttt{text-embedding-3-large}. We compute a Vendi score \citep{friedman2023vendi} on the resulting dot-product kernel and log the cumulative normalized Vendi gain when adding each new batch of generated programs to the previous curriculum. This measures program-level embedding diversity; it is not used as a reward.

Second, we embed the generated synthetic problem statement and compare it to the original target problem statement using cosine similarity. This measures text-level target relatedness and is separate from program-level novelty.

\begin{figure}[!htbp]
    \centering

    \begin{subfigure}[t]{0.49\textwidth}
        \centering
        \includegraphics[width=\linewidth]{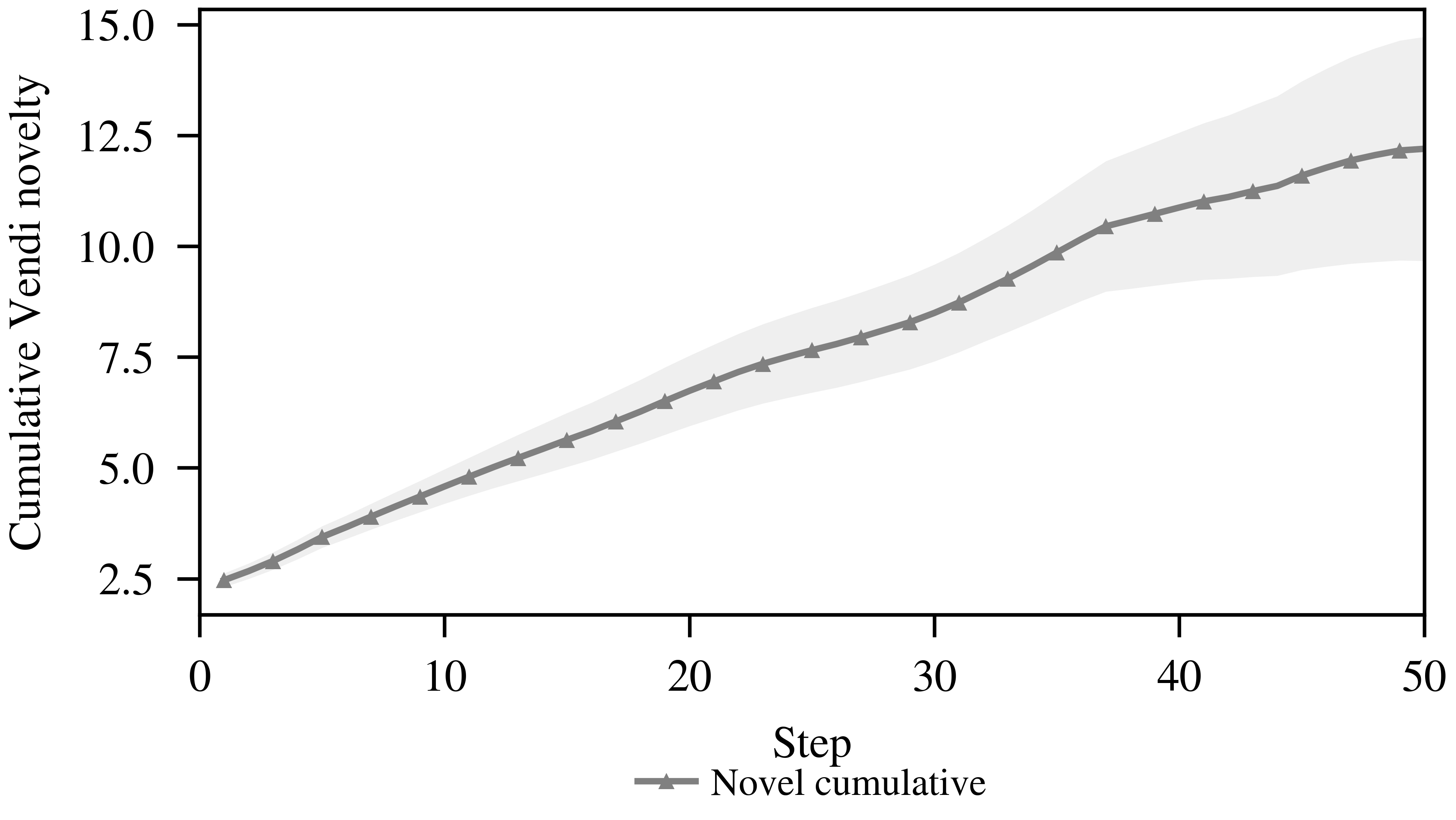}
        \caption{Cumulative normalized Vendi novelty over extracted Python programs.}
        \label{fig:vendi_novel_cumulative}
    \end{subfigure}
    \hfill
    \begin{subfigure}[t]{0.49\textwidth}
        \centering
        \includegraphics[width=\linewidth]{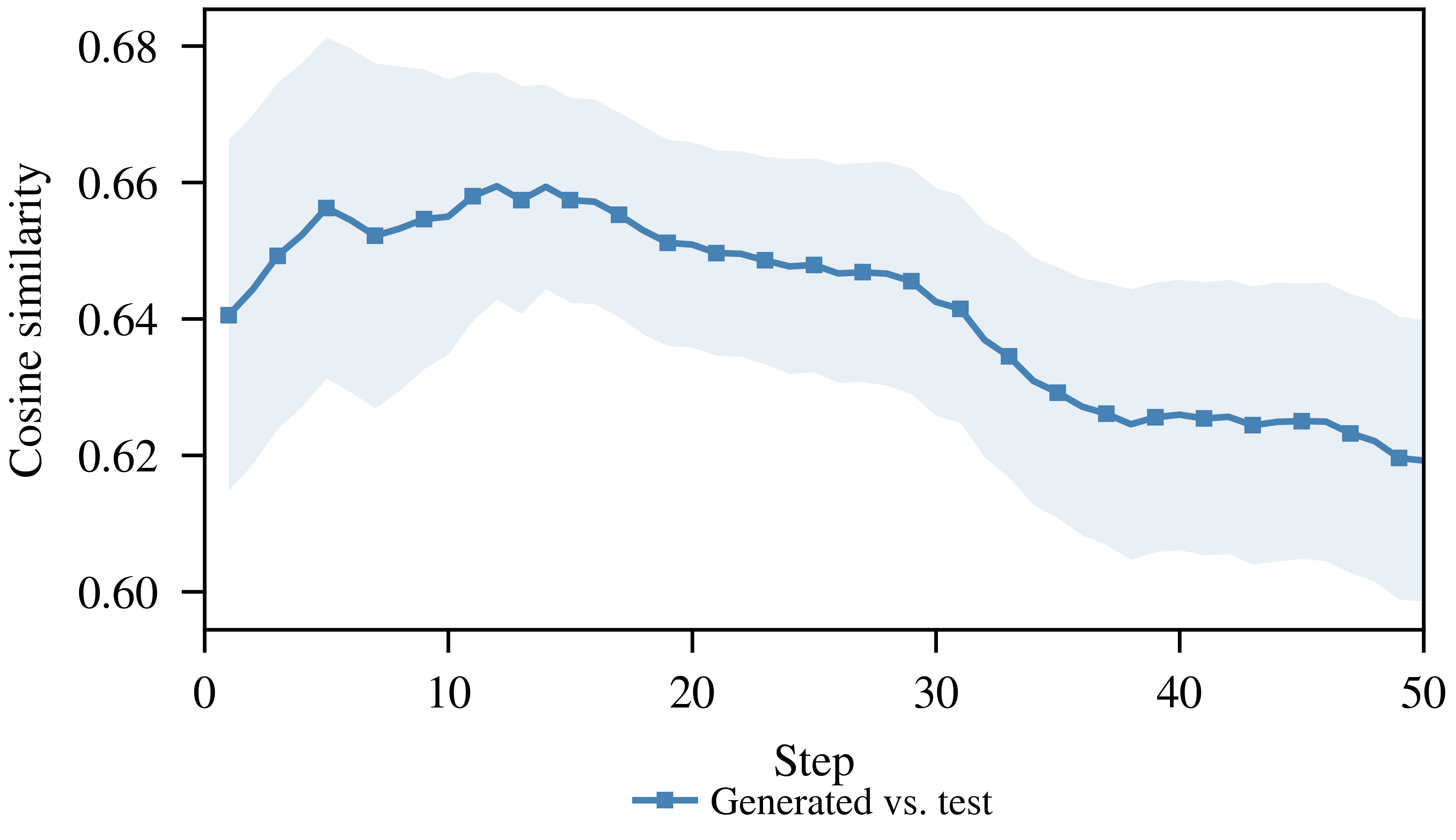}
        \caption{Cosine similarity between generated synthetic problem text and the target problem text.}
        \label{fig:cosine_subgoals_test}
    \end{subfigure}

    \caption{\textbf{Curriculum novelty and target relatedness.}
    Left: cumulative normalized Vendi gain computed from embeddings of extracted Python reference programs. Right: text-level cosine similarity between generated synthetic problem statements and the original target problem statement. Shaded bands denote $\pm 1$ SEM across target problems.}
    \label{fig:embedding_novelty_target_similarity}
\end{figure}

Figure~\ref{fig:embedding_novelty_target_similarity} shows two trends. Cumulative Vendi gain increases over training, which is inconsistent with repeatedly adding only near-identical embedded reference programs. Text-level similarity to the target peaks early and then decreases. These diagnostics do not prove semantic equivalence to the target problem; they only audit diversity and textual relatedness of the generated curriculum.

\clearpage
\section{Per-Target Results and Base-Policy Output Distributions}
\label{app:per-target-results}

This appendix expands the aggregate curves in Section~\ref{sec:main-results}. The per-target TTSP curves show how adaptation varies across problems, while the base-policy output distributions explain when majority-style methods are vulnerable to selecting the wrong functional cluster.

\subsection{TTSP Individual Results}
\label{app:ttsp-individual-results}

The aggregate curves in Figure~\ref{fig:ttt_results_aggregate_hard} show that TTSP improves quickly and remains relatively flat across verifier strengths. The per-target plots below expose the heterogeneity behind that aggregate. They also support the phase-diagram interpretation in Figure~\ref{fig:phase_diagram}: TTSP is most useful on targets with enough base model coverage to support curriculum composition, and much less reliable when the correct solution is nearly absent from the base distribution.

\begin{figure}[!htbp]
    \centering
    \includegraphics[width=0.9\textwidth]{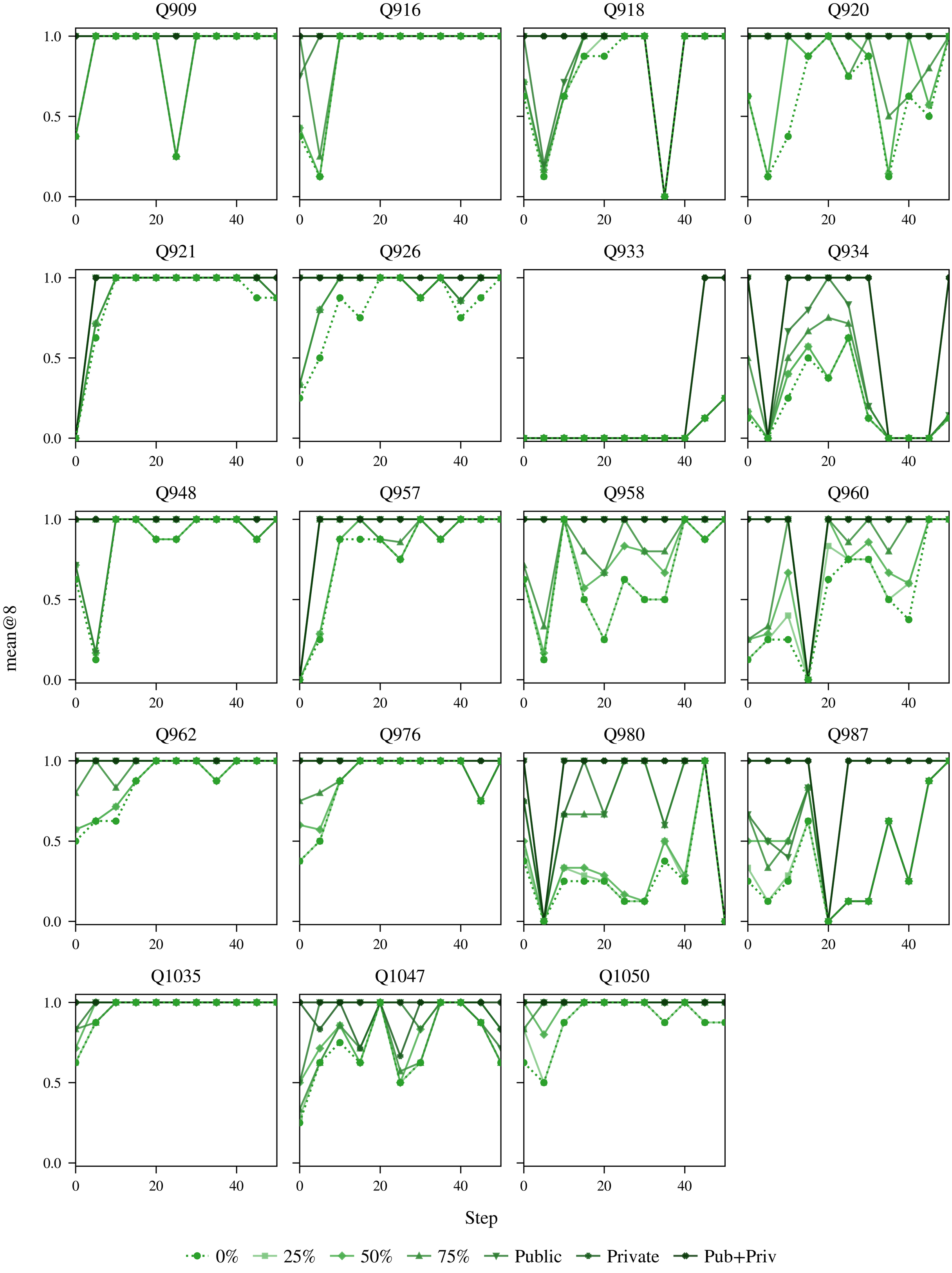}
    \caption{\textbf{Per-target TTSP training curves on MidHardLCBv6.}
    Mean@8 over GRPO steps for each individual target problem, with one panel per target and one line per evaluation-time verifier strength. Many targets show the rapid rise within $10$--$20$ steps and stable plateau seen in the aggregate curve of Figure~\ref{fig:ttt_results_aggregate_hard}.}
    \label{fig:ttt_individual_hard}
\end{figure}

\subsection{Functional Output Distributions}
\label{app:functional-output-distributions}

Section~\ref{sec:main-results} argues that FMV succeeds when the correct functional cluster is modal and fails when it is not. The plots in this appendix visualize the base policy's functional output distributions directly. Each cluster is defined by execution behavior over target problem test inputs, as in FMV-style consensus selection \citep{launer2026fmv}. These distributions explain why majority-based selection can be brittle in the consistency regime: the correct solution may be present in the sample set but still be ranked below a more common incorrect functional cluster.

\begin{figure}[!htbp]
    \centering
    \includegraphics[width=0.9\textwidth]{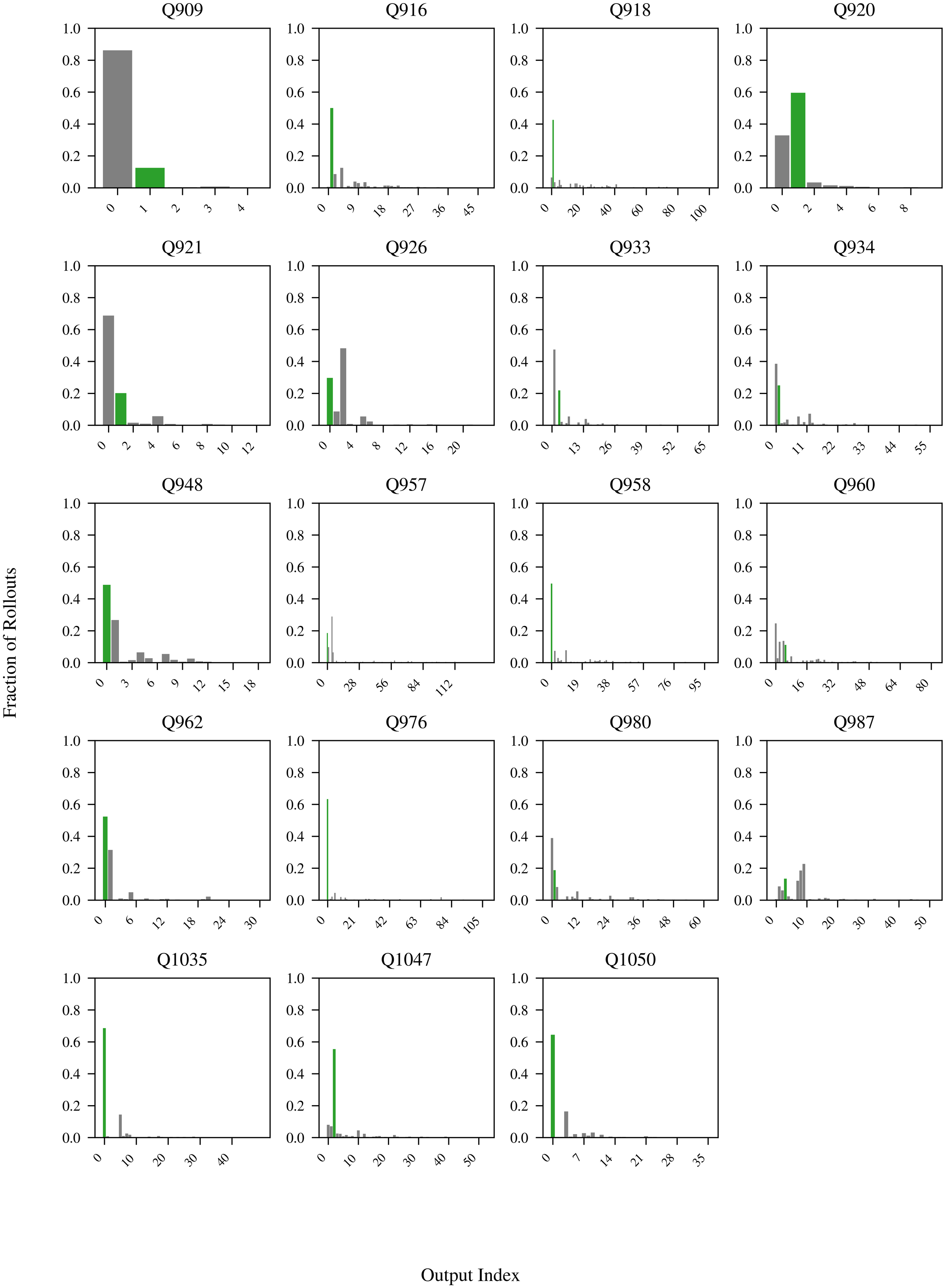}
    \caption{\textbf{Functional output clusters for MidHardLCBv6.}
    Qwen 3 8B functional output cluster distribution for each target problem in MidHardLCBv6. Each panel shows, for one target, the fraction of $512$ base rollouts whose execution traces fall into each functional cluster, with the correct cluster highlighted in green.}
    \label{fig:output_hard}
\end{figure}

\begin{figure}[!htbp]
    \centering
    \includegraphics[width=0.78\textwidth]{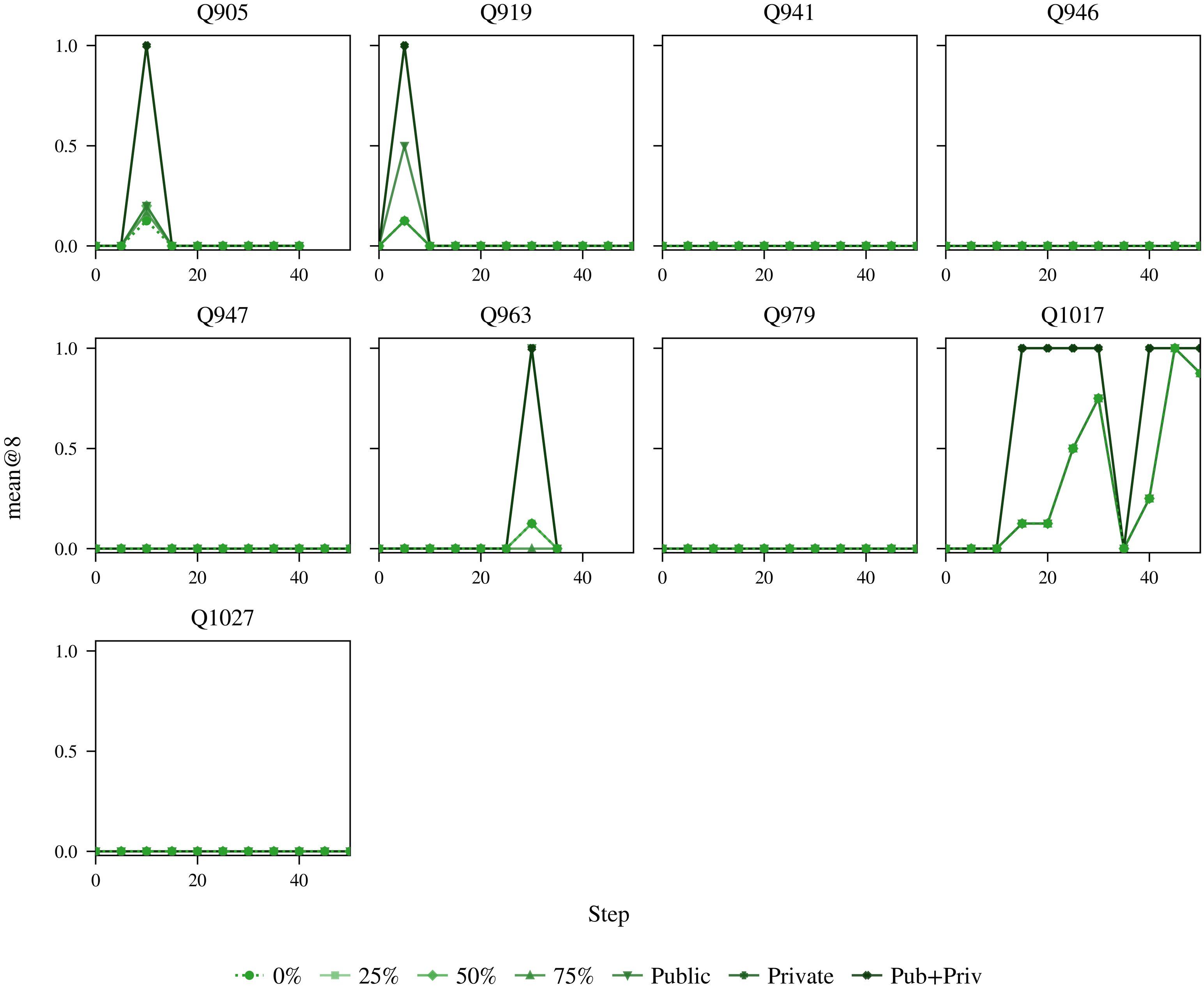}
    \caption{\textbf{Per-target TTSP training curves on VeryHardLCBv6.}
    Mean@8 over GRPO steps for each individual target problem, with one panel per target and one line per evaluation-time verifier strength. The flat or noisy curves illustrate the discovery-side failure mode discussed in Section~\ref{sec:main-results} when base coverage is below $\mathrm{mean}@512 \approx 0.1$.}% the proposer policy cannot bootstrap a curriculum that reliably composes back to the target problem.}
    \label{fig:ttt_individual_veryhard}
\end{figure}

\begin{figure}[!htbp]
    \centering
    \includegraphics[width=0.78\textwidth]{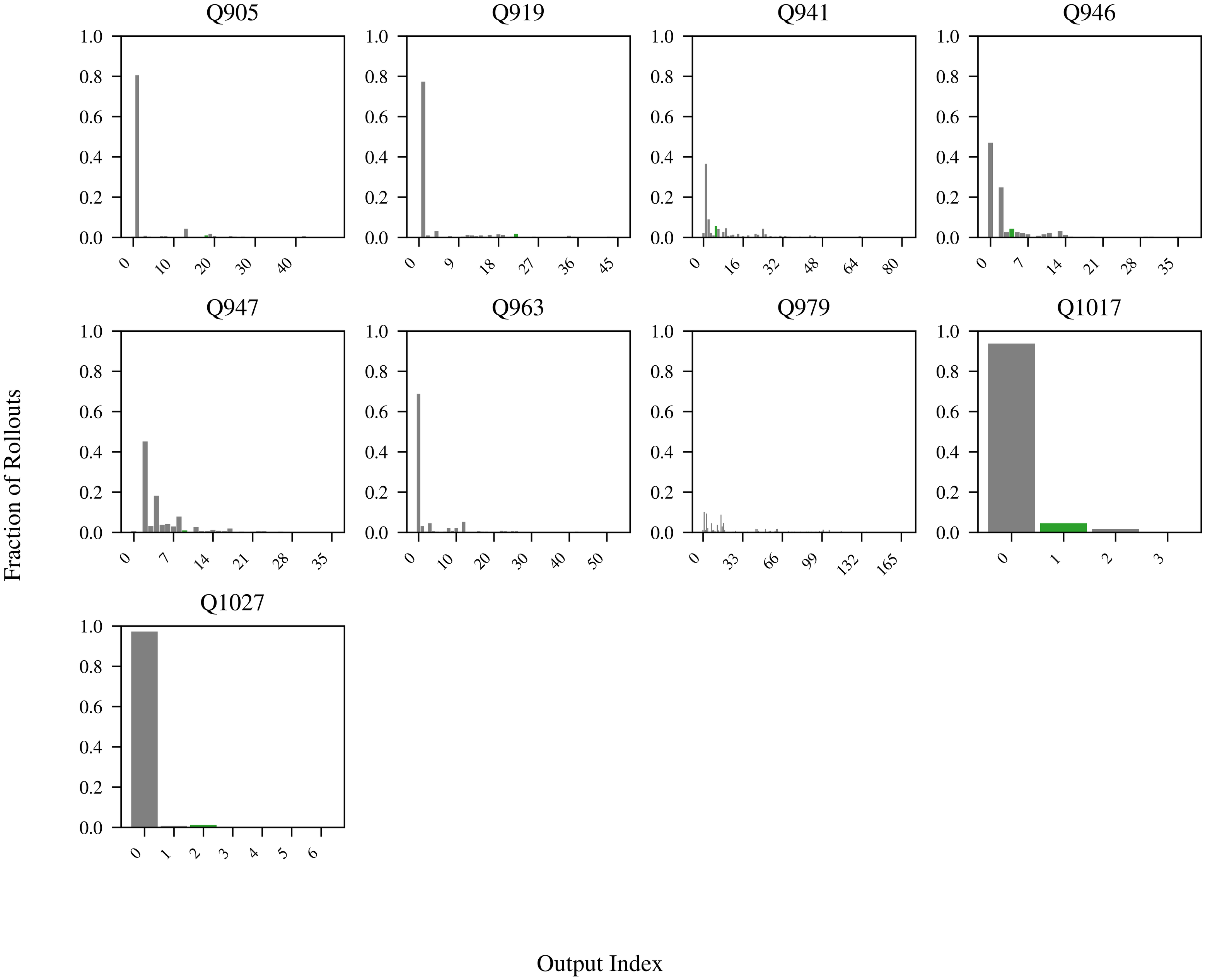}
    \caption{\textbf{Functional output clusters for VeryHardLCBv6.}
    Qwen 3 8B functional output cluster distribution for each target problem in VeryHardLCBv6. Each panel shows, for one target, the fraction of $512$ base rollouts whose execution traces fall into each functional cluster, with the correct cluster highlighted in green. The correct cluster typically carries negligible mass, matching the discovery-regime characterization in Figure~\ref{fig:ttt_delta}.}
    \label{fig:output_veryhard}
\end{figure}

\clearpage
\section{Prompt Templates}
\label{app:prompts}

The method description in Section~\ref{sec:method} abstracts TTSP into proposer generation, refinement, solver attempts, and execution-based rewards. For reproducibility, we report the exact prompt templates used to instantiate those calls. The prompts are implementation details rather than conceptual contributions, so they appear after the experimental and diagnostic appendices. They are reproduced verbatim from the final system and therefore retain the older internal naming used in the implementation: 
\texttt{target\_question} denotes the target problem $x^\star$, and 
\texttt{subgoal} denotes what the main text calls a synthetic problem. 
Placeholders such as \texttt{\{target\_question\}}, \texttt{\{context\}}, 
\texttt{\{subgoal\}}, \texttt{\{example\_input\}}, and 
\texttt{\{example\_output\}} are filled at runtime.

\subsection{Initial proposer policy prompt}
\label{app:initial_proposer_prompt}

\begin{PromptBlock}{Initial proposer prompt}
"""
## Task
You are creating a curriculum of subgoals to help a student learn the necessary skills to solve a target question.
Given the target question, propose a new subgoal and write a Python code that solves the subgoal. You must:
1. **First, think step-by-step** about what skills/concepts the student needs to learn to solve the target question.
2. **Then, propose a subgoal** that is reasonably challenging for the student and builds toward solving the hard task.
3. **Finally, write the Python ground truth code** that solves the subgoal.

**IMPORTANT: You MUST include detailed reasoning before proposing your subgoal. Responses that skip the thinking step will be rejected.**

---

## Target question
{target_question}

---

## Step 1: Reasoning (REQUIRED)
1. **Identify the core skills**: What algorithmic techniques, data structures, or mathematical concepts are relevant to the target question?
2. **Identify possible subgoals**: Based on the required algorithmic techniques, data structures, or mathematical concepts, identify and discuss three possible subgoals.

Be as specific as possible in your reasoning.

---

## Step 2: Subgoal Proposal (REQUIRED)

After your reasoning, propose a subgoal that
- **Targets relevant concepts**: The subgoal must practice concepts, knowledge and skills needed for the target question.
- **Is well-posed and complete**: The subgoal must be well-posed and complete without requiring any external context to understand.
- **Is unambiguous**: The subgoal must be as descriptive as possible so that it is completely unambiguous. Define every term in the subgoal, including all inputs and outputs and what they represent.

You must wrap your subgoal in ```subgoal``` tags.

---

## Step 3: Code Requirements (REQUIRED)

After the subgoal, write the Python code that fully solves the subgoal. Do not just write a placeholder function, write the actual code that solves the subgoal.

Your code must meet the following requirements:

- Must accept input from stdin
- Ensure the code prints a value to stdout
- Include at least one input parameter
- Make the function deterministic
- Make the snippet require state tracking across multiple data transformations, ensuring the task requires long multi step reasoning
- AVOID THE FOLLOWING:
* Random functions or variables
* Date/time operations
* I/O operations (reading files, network requests)
* Printing or logging
* Any external state
- Ensure execution completes within 10 seconds on a modern CPU
- All imports and class definitions should be at the very top of the code snippet
- The snippet should end with a print statement to stdout, only print the final result
- Do not print any example cases you make up yourself

You must wrap your code in ```python``` tags.

---

## Output Format

You must structure your response in this exact order:

[Write the detailed step-by-step reasoning. This section must be at least 4 paragraphs long.]

Then:

```subgoal
Write a function that takes a list of numbers and returns the sum of the numbers.
```

Then:

```python
arg = input()
# your code here
print(...)
```

---

## Final reminder
You must always reason step-by-step first, then propose a subgoal, then write the Python code that solves the subgoal. Write a complete python code solution that fully solves the subgoal, do not just write a placeholder function or only part of the solution.
"""
\end{PromptBlock}

\subsection{Iterative proposer policy prompt}
\label{app:next_proposer_prompt}

\begin{PromptBlock}{Iterative proposer prompt}
"""
## Task
Given the target question and past attempted subgoals, reason about what the next subgoal should be, and then propose a new subgoal, and then write a Python code that solves the subgoal.

## Target question
{target_question}

{context}

## Task description
1. Using the summary of the previous subgoals, reason about what the next subgoal should be. Carefully devise a subgoal plan: e.g., understand the previous subgoals, reason about a relevant, appropriately challenging and creative next subgoal. Isolate specific difficulties the student is having.
2. Write the subgoal. Wrap your subgoal in ```subgoal``` tags.
3. Write the solution to the subgoal in ```python``` tags. You can use the summary of previous subgoals to help guide your solution.

## Subgoal Requirements
- The subgoal must target concepts, knowledge and skills in which the student performed poorly in previous subgoals
- The subgoal must be well-posed and complete without requiring any external context to understand
- The subgoal must be challenging for the student. If all previous subgoals are too easy, suggest a subgoal that is more challenging (for example, one which gives less hints about the solution, or is more complex, or tackles a different aspect of the target question).
- The subgoal MUST be at least slightly different from previous subgoals
- The subgoal must be relevant to the target question
- Define every term in the subgoal, including all the inputs and outputs and what they represent
- The subgoal must be as descriptive as possible so that it is completely unambiguous

### Subgoal format:
- Wrap your subgoal in ```subgoal``` tags.
- For example:
```subgoal
Write a function that takes a list of numbers and returns the sum of the numbers.
```

### Code Requirements:
- Must accept input from stdin
- Ensure the code prints a value to stdout
- Include at least one input parameter
- Make the function deterministic
- Make the snippet require state tracking across multiple data transformations, ensuring the task requires long multi step reasoning
- AVOID THE FOLLOWING:
* Random functions or variables
* Date/time operations
* I/O operations (reading files, network requests)
* Printing or logging
* Any external state
- Ensure execution completes within 10 seconds on a modern CPU
- All imports and class definitions should be at the very top of the code snippet
- The snippet should end with a print statement to stdout, only print the final result
- Do not print any example cases you make up yourself

### Formatting:
- Format your code with:
```python
arg = input()
# your code here
print(...)
```

## Rules
- Be succinct!
- Don't overanalyze. Do not analyze every task.
- Do not worry about being too descriptive in the subgoal, as long as it is unambiguous and complete.
- If all previous subgoals are too easy, suggest a subgoal that tackles a different aspect of the target question, or is more challenging.

YOU MUST ALWAYS WRITE BOTH A SUBGOAL AND A PYTHON CODE SOLUTION.
"""
\end{PromptBlock}

\subsection{Solver policy prompt}
\label{app:solver policy_prompt}

\begin{PromptBlock}{Solver prompt}
"""
## Task
You are solving a programming subgoal. Given a subgoal description and examples, you must reason through the solution step-by-step and then write Python code that solves it.

**IMPORTANT: You MUST include detailed reasoning before writing your code. Responses that skip the thinking step will be rejected.**

---

## Subgoal
{subgoal}

### Example input:
{example_input}

### Example output:
{example_output}

---

## Step 1: Reasoning (REQUIRED)

Before writing code, think step-by-step:
1. **Understand the subgoal**: What is the input? What is the expected output? What are the constraints?
2. **Identify the approach**: What algorithm or technique will you use? Why is it appropriate?
3. **Plan the implementation**: What are the key steps in your solution?

---

## Step 2: Solution (REQUIRED)

After your reasoning, write clean Python code that solves the subgoal.

You may use any appropriate techniques including:
- Data structures: trees, heaps, stacks, queues, graphs, hash maps, etc.
- Algorithms: dynamic programming, recursion, divide and conquer, greedy, backtracking, BFS/DFS, binary search, etc.

### Program format:
- Wrap your Python code in ```python``` tags.
- Must accept input from stdin
- Ensure your code prints a value to stdout
- Make the function deterministic
- AVOID THE FOLLOWING:
* Random functions or variables
* Date/time operations
* I/O operations (reading files, network requests)
* Printing or logging
* Any external state
- Ensure execution completes within 10 seconds on a modern CPU
- All imports and class definitions should be at the very top of the code snippet
- Use imports if it makes it easier to solve the subgoal
- The snippet should end with a print statement to stdout, only print the final result
- Your code will be tested on hidden test cases with varied inputs
- Your code must compute the answer, not assume it
- For example:
```python
t = input()
# code logic here
print(result)
```

---

## Output Format

You must structure your response in this exact order:

### Reasoning
Analyze the subgoal, identify the algorithm/approach, and plan your implementation.

### Solution
```python
t = input()
# code logic here
print(result)
```

Do NOT use if __name__ == "__main__" in your code.
Include comments explaining your approach. Remember to wrap your final code block in ```python and ``` tags."""
\end{PromptBlock}

\subsection{Refinement / labeling prompt}
\label{app:refinement_prompt}

\begin{PromptBlock}{Refinement / labeling prompt}
"""
{ground_truth}

## Task
Given the above Python code, you must:
1. Analyze the code thoroughly
2. Create a competitive programming subgoal where the Python code is the solution
3. Create {num_inputs} sets of example inputs for the code

**IMPORTANT: You MUST include detailed reasoning before writing your subgoal and test cases. Responses that skip the thinking steps will be rejected.**

---

## Step 1: Code Analysis (REQUIRED)

Before creating anything, analyze the code step-by-step:
1. **Input parsing**: What inputs does the code expect? What are their types and formats (integers, strings, lists, etc.)?
2. **Core logic**: What algorithm or approach does the code implement? What is its purpose?
3. **Output**: What does the code produce? What is the output format?
4. **Constraints**: What are the implicit constraints (e.g., input ranges, sizes)?
5. **Edge cases**: What boundary conditions does the code handle?

---

## Step 2: Subgoal Design (REQUIRED)

Design each component of the subgoal statement:

1. **Subgoal narrative**: Create a clear scenario or abstract description that motivates the subgoal. What task is the solver asked to accomplish?

2. **Input specification**: Define the exact input format:
   - How many lines of input?
   - What does each line contain?
   - What are the data types (integers, strings, etc.)?
   - Are there delimiters (spaces, commas)?

3. **Output specification**: Define the exact output format:
   - What should be printed?
   - What format (single number, space-separated values, etc.)?

4. **Constraints**: Specify bounds that match the code's assumptions:
   - Value ranges (e.g., 1 \leq n \leq 10^5)
   - String lengths
   - Array sizes
   - Time/memory limits if relevant

---

## Step 3: Test Case Design (REQUIRED)

Plan your {num_inputs} test inputs to ensure good coverage:
1. Include at least one simple/basic case
2. Include edge cases (e.g., minimum values, boundary conditions)
3. Include cases that exercise different code paths
4. Ensure all inputs are valid and will run successfully with the code

---

## Output Format

You must structure your response in this exact order:

### Code Analysis
Describe your understanding of the code's inputs, logic, and outputs.

### Subgoal Design
Detail your decisions for: narrative, input format, output format, constraints, and examples.

### Test Case Design
Explain your strategy for creating diverse, valid test inputs.

### Subgoal Statement
```subgoal
[Your competitive programming subgoal statement here. Must include:
- Clear subgoal description/narrative
- Input format specification
- Output format specification  
- Constraints
- At least one example input/output

Do NOT leak any implementation details or algorithm hints.]
```

### Test Cases
Provide exactly {num_inputs} sets of inputs:
```input
arg1\\narg2\\n...
```
```input
arg1\\narg2\\n...
```

---

## Important Rules
- The subgoal statement must EXACTLY and correctly describe what the Python code does
- Do NOT leak any information about the Python solution in the subgoal statement
- Do NOT mention specific algorithms, data structures, or implementation hints
- Each set of inputs must be enough to run the Python code once
- Format multiple arguments delimited by \\n
- Do NOT add quotes around arguments. Write values exactly as they would appear in stdin
- Do NOT provide explanations of test cases after the test inputs
"""
\end{PromptBlock}

\end{document}